\documentclass[pdflatex,sn-mathphys-ay]{sn-jnl}

\usepackage{graphicx}%
\usepackage{multirow}%
\usepackage{amsmath,amssymb,amsfonts}%
\usepackage{amsthm}%
\usepackage{mathrsfs}%
\usepackage[title]{appendix}%
\usepackage{xcolor}%
\usepackage{textcomp}%
\usepackage{manyfoot}%
\usepackage{booktabs}%
\usepackage{algorithm}%
\usepackage{algorithmicx}%
\usepackage{algpseudocode}%
\usepackage{listings}%
\usepackage{hyperref}

\theoremstyle{thmstyleone}%
\theoremstyle{thmstyletwo}%

\theoremstyle{thmstylethree}%
\newcommand{\new}[1]{\textcolor{black}{#1}}
\newcommand{\ext}[1]{\textcolor{black}{#1}}
\newcommand{\added}[1]{\textcolor{black}{#1}}

\begin{document}

\title{Structured Sentiment Analysis as Transition-based  Dependency \added{Graph} Parsing}

\author{\fnm{Daniel} \sur{Fern\'{a}ndez-Gonz\'{a}lez}}\email{danifg@uvigo.es}

\affil{\orgname{Universidade de Vigo}, \orgdiv{Departamento de Informática}, \orgaddress{\street{As Lagoas s/n}, \city{Ourense}, \postcode{32004}, \country{Spain}}}


\abstract{Structured sentiment analysis (SSA) aims to automatically extract people's opinions from a text in natural language and adequately represent that information in a graph structure. One of the most accurate methods for performing SSA was recently proposed and consists of approaching it as a dependency \added{graph} parsing task. Although we can find in the literature how transition-based algorithms excel in \added{different dependency graph parsing tasks} in terms of accuracy and efficiency, all proposed attempts to tackle SSA following that approach were based on graph-based models. In this article, we present the first transition-based method to address SSA as dependency \added{graph} parsing. Specifically, we design a transition system that processes the input text in a left-to-right pass, incrementally generating the graph structure containing all identified opinions. To effectively implement our final transition-based model, we resort to a Pointer Network architecture as a backbone. From an extensive evaluation, we demonstrate that our model offers the best performance to date in practically all cases among prior dependency-based methods, and surpasses recent task-specific techniques on the most challenging datasets. We additionally include an in-depth analysis and empirically prove that the 
\added{average-case} time complexity of our approach is quadratic in the sentence length, being more efficient than top-performing graph-based parsers.
\\

\textbf{This version of the article has been accepted for publication in Artificial Intelligence Review after peer review, but is not the Version of Record and does not reflect post-acceptance improvements, or any corrections. The Version of Record is available online at: \url{https://doi.org/10.1007/s10462-025-11463-9}.}
}


\maketitle

\section{Introduction}
With the dramatic growth of user-generated data 
on the Web and social media platforms,
\textit{sentiment
 analysis}
(SA) \citep{SA}  has become an 
emerging and challenging research topic in natural language processing (NLP). It focuses on developing techniques to automatically analyze the sentiment and opinions expressed by individuals 
in various text sources, such as reviews, customer feedback and social media posts. 
In the last decade, SA has received increasing attention due to 
its real-world applications. Among them, we can highlight the processing of people's opinions on social networks 
\citep{8959677,Wang2020,10632865} 
in order to identify fake news 
\citep{Alonso2021}
or monitor political opinions \citep{COTELO2016}; as well as providing valuable insights into customer sentiment and, as a consequence, helping companies to improve their products and services 
\citep{Pu2019,10016285}.


Early research on SA focused on a coarse-grained analysis, consisting of merely determining the overall sentiment polarity (i.e., \textit{positive}, \textit{negative},  or  \textit{neutral}) of a whole sentence or document \citep{dong-etal-2014-adaptive,nguyen-shirai-2015-phrasernn}. However, the tremendous progress in NLP techniques
allowed to perform a more fine-grained analysis such as: \new{\textit{aspect-based sentiment analysis} (ABSA), which identifies the sentiment polarities of different aspects or targets within the same sentence 
\citep{sun-etal-2019-aspect,tang-etal-2020-dependency,apostol2023atesabaertheterogeneousensemblelearning};
\textit{opinion role labeling},
which captures the text fragment expressing the opinion along with the entity that was targeted and the entity holding that opinion 
\citep{Zhang2019,xia-etal-2021-unified}; or \textit{aspect sentiment quad prediction}, which seeks to identify the target entity, the relevant aspect, the sentiment polarity, and the opinion terms that express the sentiment \citep{zhang-etal-2021-aspect-sentiment,zhu-etal-2024-pinpointing,su-etal-2025-unified}.}

\begin{figure}[h]
\centering

\includegraphics[width=0.8\textwidth]{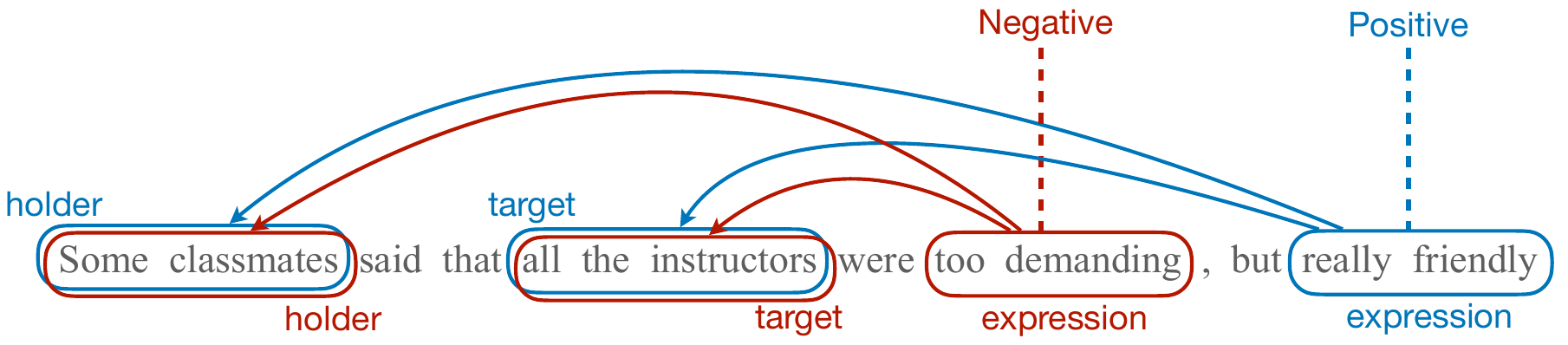}%
\\(a)\\
\label{fig_sg}
\vspace{0.1in}

\includegraphics[width=\textwidth]{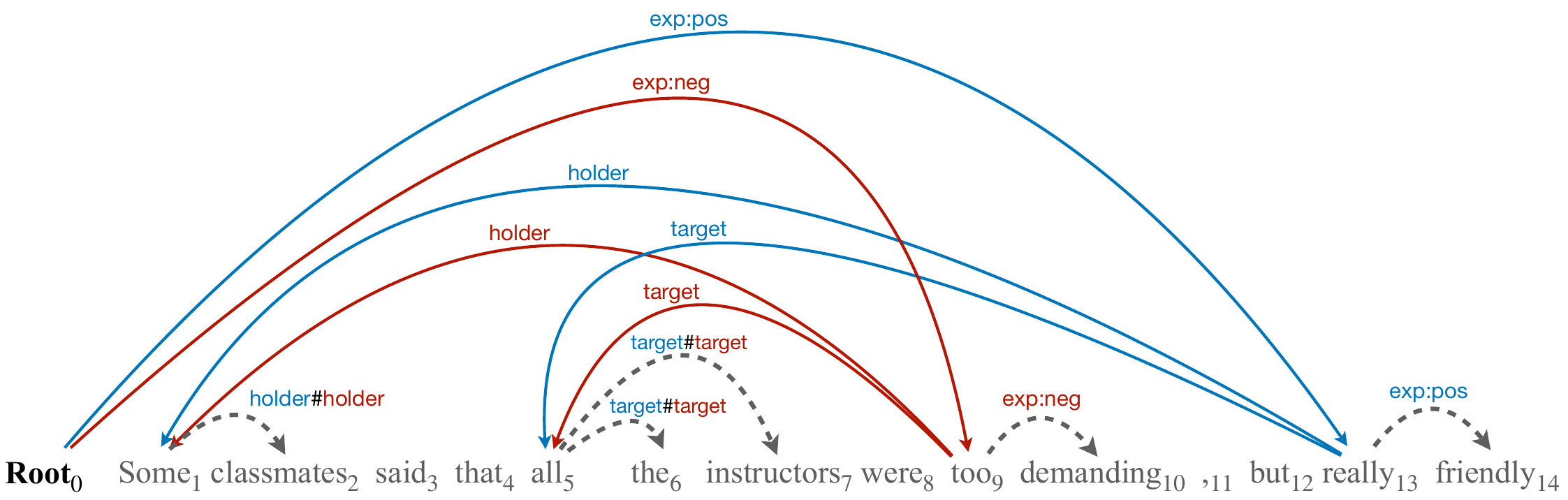}%
\\(b)\\
\label{fig_first}
\vspace{0.1in}

\includegraphics[width=\textwidth]{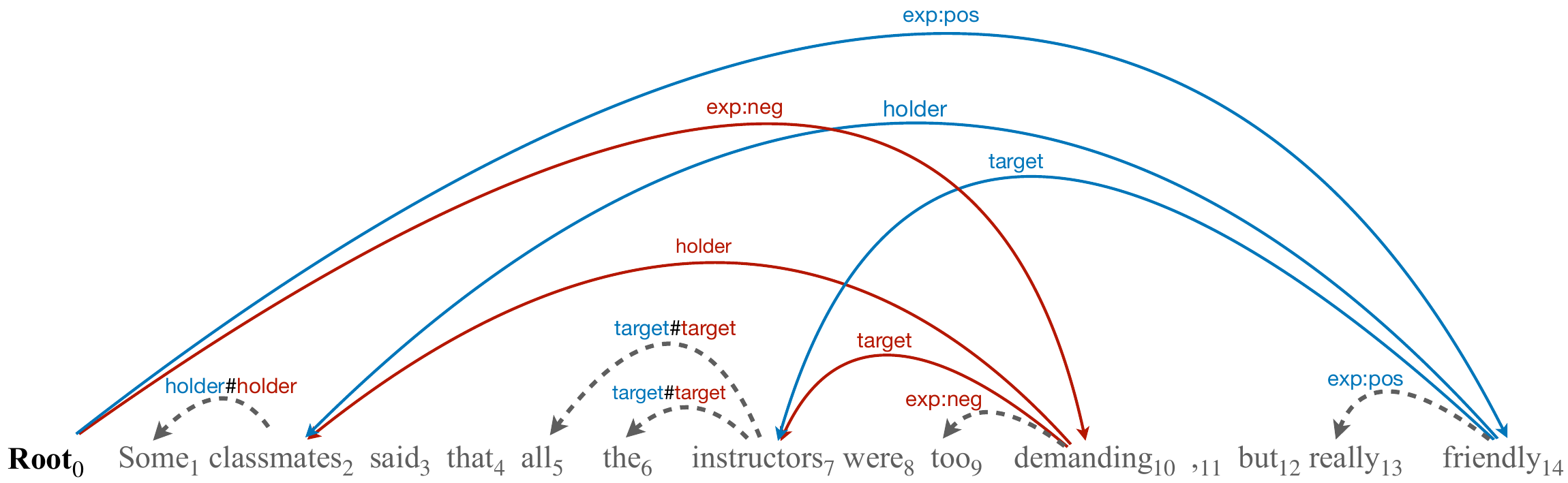}%
\\(c)\\
\label{fig_final}
\vspace{0.1in}
\caption{A sentiment graph (a) is encoded into bi-lexical dependency graphs following the head-first (b) or head-final (c) strategies.}
\label{fig_1}
\end{figure}

\new{\cite{Barnes2021} further expanded fine-grained SA tasks by introducing \textit{structured sentiment analysis} (SSA). This is} a more complex task that, given an input text, aims to predict a richer annotation: a \textit{sentiment graph} that fully represents all the opinions along with their arguments
 and sentiment polarity. An example of a sentiment graph is depicted in Figure~\ref{fig_sg}(a), where, for instance, the \textit{opinion expression} ``too demanding'' describes how the opinion is given, the \textit{opinion holder} ``Some classmates'' represents who holds the opinion statement, the \textit{opinion target} ``all the instructors'' identifies to whom the opinion is addressed, and the \textit{sentiment polarity} ``Negative'' summarizes what is the polarity of that opinion. We can also see as the opinion expression is the root node of the resulting sentiment graph, being connected to its respective holder and target arguments by directed arcs. Finally, we additionally present in that example a positive opinion expressed by the text fragment ``really friendly'', which shares the same holder and target as the negative opinion but is represented by a separate graph structure. 

SSA can be divided into five sub-tasks: \textit{i)} opinion expression extraction, \textit{ii)} opinion holder identification, \textit{iii)} opinion target identification, \textit{iv)} predicting the relationship between these entities, and, finally, \textit{v)} assigning the sentiment polarity of the overall opinion. 
\new{One or several of these sub-tasks have already been addressed by existing methods, which can be integrated to achieve SSA; however, \cite{Barnes2021} proposed to fully resolve the SSA task with a single end-to-end model, arguing that a pipeline framework could be counterproductive. In particular, they define a conversion method to cast SSA as a \textit{dependency \added{graph} parsing} problem.}


\added{Dependency graph parsing involves identifying \textit{bi-lexical} relationships (known as dependencies) that link a \textit{head} word to its dependent, forming a directed dependency graph. In contrast to sentiment graphs (\textit{n}-lexical structures where nodes may represent entire text chunks), dependency graph nodes correspond exclusively to individual words. \cite{Barnes2021} proposed leveraging dependency graphs to encode sentiment graphs. This approach involves transforming sentiment graph nodes into a bi-lexical \textit{dependency-based} framework.}\footnote{\ext{Please note that, although dependency graphs are typically used to represent syntactic information, in this work they are employed to encode sentiment graph nodes and to capture the relationships between the elements that constitute an opinion within the sentiment graph.}} In Figures~\ref{fig_first}(b) and \ref{fig_final}(c), we present two different dependency-based representations of the sentiment graph in Figure~\ref{fig_sg}(a). 

\new{Within this framework, all opinion components can be jointly identified by predicting a single dependency graph. In particular, \citet{Barnes2021} resorted to the \textit{graph-based} parser introduced by \citet{dozat-manning-2018-simpler}, which independently scores all potential dependencies and conducts an exhaustive search to find the highest-scoring graph in \( O(n^2) \) time (where \( n \) is the length of the input sentence). Once a well-formed sentiment graph is extracted from the resulting dependency structure, the entire SSA task is effectively performed by a single parser.}


While this \textit{dependency-based} reduction to intermediate representations is lossy due to an inherent ambiguity in the encoding scheme \citep{Samuel2022}, 
this technique obtained strong improvements in SSA and managed to notably outperform
previous 
state-of-the-art 
methods in other fine-grained SA tasks, such as 
ABSA \citep{he-etal-2019-interactive,chen-qian-2020-relation}. 
In fact, 
its success motivated other research works \citep{Peng2021} to further improve the original model by adopting 
a more advanced graph-based parser, obtaining notable improvements in accuracy but also increasing 
the overall runtime complexity to $O(n^3)$.

\new{To date, all existing approaches that formulate SSA as a dependency \added{graph} parsing task have relied on graph-based parsers. However, a well-established alternative exists in the form of \textit{transition-based} algorithms, which have been extensively studied for \added{different dependency graph parsing tasks}. Unlike graph-based methods, they process an input sentence from left to right by locally predicting a sequence of actions that incrementally generates a dependency graph. The resulting number of actions is linear 
with respect to the sentence length, being in general more efficient than graph-based models. In fact,
the ongoing advancements in transition-based techniques have established them as some of the most accurate and efficient approaches in \added{different NLP tasks 
 \citep{fei-end-SRL,Wu2022ORL,FerGomKBS2022}.}
 Despite this advantage, no attempt has been made to approach SSA as transition-based dependency \added{graph} parsing.}

Far from the mainstream,
we present the first work addressing SSA with a transition-based method. 
\new{We define a transition system that builds 
a dependency-based representation
by traversing the sentence from left to right, connecting each word to zero, one, or multiple head words. 
As dependency-based encoding strategies, we explore all methods introduced in \citep{Barnes2021}, which, after decoding, yield a complete sentiment graph.} 
Following recent state-of-the-art works \citep{fernandez-gonzalez-gomez-rodriguez-2020-transition,FerGomKBS2022}, we implement a Pointer Network \citep{Vinyals15} as the backbone of our transition-based model and exploit it to accurately predict the sequence of actions required by the transition system. 

The resulting approach is extensively evaluated on five standard benchmarks in four languages (English, Norwegian, Basque, and Catalan), including NoReC$_{\text{Fine}}$ \citep{ovrelid-etal-2020-fine}, MultiB$_{\text{EU}}$, MultiB$_{\text{CA}}$ \citep{barnes-etal-2018-multibooked}, MPQA \citep{Wiebe2005AnnotatingEO} and DS$_{\text{Unis}}$ \citep{toprak-etal-2010-sentence}. Experiments show that our method outperforms the original work by \cite{Barnes2021} in all benchmarks and it is on par with the best-performing methods recently proposed, including among them task-specific techniques that do not require an intermediate dependency-based encoding. This demonstrates that a transition-based approach is a promising option for solving the SSA problem as dependency \added{graph} parsing. 
 
 Lastly, 
 we further investigate the leverage of syntactic information for encoding sentiment graphs into dependencies; as well as, prove that our end-to-end technique processes all datasets in $O(n^2)$ time in practice, resulting in one of the most efficient approaches for tackling SSA.

In summary, we advance the research in SSA with the following contributions:
\begin{itemize}
    \item We develop the first approach for performing end-to-end SSA as transition-based dependency \added{graph} parsing.
    \item Our model is robust and achieves strong results across all benchmarks, improving over graph-based baselines in 
    four out of five datasets and obtaining comparable results to task-specific approaches in the most challenging benchmarks.
    \item We test our approach not only on the two available dependency-based representations \citep{Barnes2021},
    but additionally leverage syntactic information in the encoding process to study its impact on the final accuracy.
    \item We empirically prove that the proposed transition-based technique processes all benchmarks in $O(n^2)$ time in practice, being more efficient than the best-performing graph-based model ($O(n^3)$).
    \item \new{Our system's source code is publicly available at \url{https://github.com/danifg/SSAPointer}.}
\end{itemize}

The remainder of this article is organized as follows:
In Section~\ref{sec:related}, we introduce previous efforts in end-to-end SSA. Section~\ref{sec:model} 
describes the transition system developed for full SSA and
details the neural architecture exploited for implementing the transition-based parser. In Section~\ref{sec:experiments}, we present an extensive evaluation of our model on standard benchmarks, discuss the experimental results, study in depth the contribution of each component, investigate the impact of syntax information in the dependency-based encoding, and analyze the overall runtime complexity of our method in practice.  
Lastly, Section~\ref{sec:conclusions} presents final conclusions.

\section{Related Work}
\label{sec:related}

Since \cite{Barnes2021} introduced SSA,
several efforts have tried to fully address that task following an end-to-end framework. In that initial work, they propose to approach SSA as dependency \added{graph} parsing by transforming original sentiment graphs into bi-lexical dependency-based representations. Then, they directly apply the model by \cite{dozat-manning-2018-simpler}: a graph-based parser that processes any sentence with length  $n$  in $O(n^2)$ by scoring each possible dependency edge and outputting the high-scoring graph. Subsequently, \cite{Peng2021} adopted the same encoding strategy and extended the original work by adding a sparse fuzzy attention mechanism to deal with the sparseness of dependency arcs and by exploiting a second-order graph-based parser. In particular, they adopted the parser developed by \cite{wang-etal-2019-second}. This model leverages second-order information (such as sibling, co-parent and grandparent relations) and scores sets of arcs to compute the highest-scoring dependency graph in $O(n^3)$. While less efficient, the approach developed by \cite{Peng2021} delivered substantial gains in accuracy with respect to the original graph-based model.

\added{In contrast to earlier efforts that utilized standalone dependency graph parsers, recent studies have shifted toward addressing full SSA by developing more sophisticated, task-specific models capable of directly generating sentiment graphs.}
\cite{Shi2022} opted for extracting
different dependency-based features
for representing nodes and edges from the original sentiment graph during training and combining them by means of \textit{graph attention networks} \citep{GAT}. Then, during decoding, they 
apply a three-stage procedure: 
they first identify spans, subsequently 
attach each holder and target node to their expression node, and, finally,
combine holder-expression and target-expression pairs 
to produce a valid sentiment graph. \cite{Zhai2023} relied on a complex dependency-based representation of the original sentiment graph, which is transformed into a \textit{unified 2D table-filling scheme}. Then, a \textit{bi-axial attention} module \citep{Wang2020biaxial} is implemented to effectively identify entities and predict relations between them to finally build a sentiment graph. \cite{Zhou2024} proposed to represent sentiment graphs as \textit{latent dependency graphs}, where nodes of the sentiment graph are treated as latent subgraphs. Then, a two-stage parsing method is applied: a parser is first employed for generating the subgraphs that represent expression nodes, and, after that, a second parser is used for building the structure of holder and target nodes (based on the expression nodes previously identified). Specifically, they implemented two jointly-trained parsers (similar to the model by \cite{wang-etal-2019-second}) enhanced with a \textit{constrained TreeCRF} \citep{zhang-etal-2022-semantic} to explicitly model latent structures.
\cite{Samuel2022} adopted the text-to-graph model developed by \cite{samuel-straka-2020-ufal} and augmented it with fine-tuned contextualized embeddings from XLM-R \citep{conneau-etal-2020-unsupervised}. Their method is then exploited for directly building edges that connect text spans.  And, finally, \cite{Xu2022} presented a complex transition-based approach equipped with: \textit{graph convolutional networks} \citep{marcheggiani-titov-2017-encoding} for encoding high-order features, a specific component exclusively trained for detecting span boundaries, and two separate classifiers for argument labeling and polarity identification.
Despite achieving the best accuracies to date,
these task-specific approaches are notably less 
efficient and more cost-intensive than methods that simply resort to a standard parser.

\new{Finally, transition-based models have been successfully applied in a wide range of NLP parsing tasks, including \textit{dependency parsing} \citep{fernandez-gonzalez-gomez-rodriguez-2019-left}, \textit{constituency parsing} \citep{attachjuxtapose}, \textit{semantic dependency parsing} \citep{fernandez-gonzalez-gomez-rodriguez-2020-transition}, generating \textit{abstract meaning representations} \citep{fernandez-astudillo-etal-2020-transition}, \textit{semantic role labeling} \citep{fei-end-SRL,FERNANDEZGONZALEZ2023110127} or \textit{opinion role labeling} \citep{Wu2022ORL}.} In this research work, we further explore transition-based algorithms to solve SSA as dependency \added{graph} parsing and compare them with 
prior graph-based alternatives. Unlike the complex transition-based method by \cite{Xu2022} that directly produces sentiment graphs, our model relies on the encoding scheme introduced by \cite{Barnes2021} to approach SSA more efficiently.
Despite being notably simpler, our transition-based model manages to not only outperform dependency-based baselines but also 
surpass the task-specific method by \cite{Xu2022} in the majority of benchmarks.



\section{Methodology}
\label{sec:model}

This section first describes the dependency-based encoding applied to cast SSA as dependency \added{graph} parsing, then introduces the transition system specifically designed for generating dependency-based representations, and, lastly, details the neural architecture developed to implement the final transition-based parser.

\subsection{Dependency-based Encoding}
SSA consists of building a sentiment graph (as the one depicted in Figure~\ref{fig_sg}(a)) that represents all the opinions present in a given sentence. This structure is a directed graph that contains all the elements necessary for defining each opinion $O_i=(e,t,h,p)$ in the text, where $e$ is the sentiment expression with a polarity $p$ expressed by the holder $h$ towards the target $t$. In particular, the sentiment expression, target and holder are represented by labeled graph nodes (which can span over multiple tokens) and unlabeled edges that connect each sentiment expression to its respective arguments. Holders and targets are optional and, therefore, an opinion can be represented without that information. Lastly, the polarity attribute is assigned to the sentiment expression node. The resulting structure can have multiple roots and be non-connected if the sentence contains two or more opinions. And we can also find as some tokens can be left out of the graph, since they do not belong to any identified opinion.

 In order to perform end-to-end SSA as dependency \added{graph} parsing, sentiment graphs are transformed into bi-lexical dependency graphs for training an independent parser. Then, during decoding, valid sentiment graphs are recovered from predicted dependency graphs. To implement that framework, we follow the work by \cite{Barnes2021} and adopt two different encoding schemes:
\begin{itemize}
    \item \textbf{Head-first:} Every node from the original sentiment graph is converted into a dependency-based structure by choosing the first token as \textit{head} (or \textit{parent}) and attaching the remaining tokens within the node span as \textit{dependents}.
    The resulting dependency arcs are labeled with the original node tag for holder and target nodes (i.e., \texttt{holder} or \texttt{target} labels) and, for expression nodes, its polarity attribute is additionally encoded in the resulting dependency label (i.e., \texttt{exp:pos}, \texttt{exp:neg} or \texttt{exp:neu} tags). Then, the head token from holder and target dependency-based representations are attached (and respectively labeled with \texttt{holder} or \texttt{target} tags) to the head identified in the expression node. Finally, the head token from the expression node becomes the root of the whole structure. 
    \item \textbf{Head-final:} This encoding scheme follows a similar strategy, but, instead of using the first token of each node span as head, it relies on the last token. Please note that in the case that all nodes from the sentiment graph were composed of a single token, the resulting dependency graph would be the same regardless of the encoding strategy utilized. 
\end{itemize}

Irrespective of the used encoding technique, the resulting labeled dependency graph is not necessarily acyclic and, unlike the original sentiment graph, it is single-rooted:
all root tokens (i.e., heads from dependency-based representations of expression nodes) are attached to an artificial root node (added at the beginning of the sentence) and the resulting dependency arcs are labeled with 
the sentiment polarity (i.e., \texttt{exp:pos}, \texttt{exp:neg} or \texttt{exp:neu} tags). In Figures~\ref{fig_first}(b) and ~\ref{fig_final}(c), we present the dependency graphs obtained from converting the sentiment graph in Figure~\ref{fig_sg}(a) following the \textit{head-first} and \textit{head-final} strategies, respectively. Dependency arcs depicted with discontinuous arrows represent the dependency-based encoding of nodes from the sentiment graph. 

It can be noticed in Figure~\ref{fig_first}(a) as, although both opinions share the same holder and target arguments, we have separate holder and target nodes in the original sentiment graph and, therefore, they have to be individually encoded for each opinion. For instance, (if we use the head-first encoding strategy) each holder node ``Some classmates'' should be encoded by building a dependency arc from head ``Some'' to dependent ``classmates'', and labeling it as \texttt{holder}. We then collapse both edges into a single one and tag the resulting arc as ``holder\#holder'', as shown in Figure~\ref{fig_first}(b). This representation is required by our transition-based algorithm to handle multiple dependency arcs between the same head and dependent. In the post-processing step, two different dependency arcs will be recovered before undertaking the conversion from the dependency-based representation to the final sentiment graph. The same procedure is applied to encode target nodes in the example presented in Figure~\ref{fig_1}(a). Please note that our encoding scheme slightly differs from the one proposed by \cite{Barnes2021}, since, in a similar scenario, their approach
 would merely keep one arc labeled as  \texttt{holder} between the head ``Some'' and the dependent ``classmates'' (discarding multiple arcs between the same head and dependent). \new{As a result, their model is unable to reconstruct the full sentiment graph after deconversion. However, their graph-based parser may benefit from a reduced label vocabulary, particularly given that arcs representing multiple dependencies are infrequent in the data.}

Therefore, given an input sentence $X = w_0, w_1, \dots, w_n$ with the artificial root node $w_0$, our parser has to build a dependency graph $G$ represented as a set of labeled dependency arcs: $G \subseteq W \times W \times L$, where $W$ is the set of input tokens ($W = \{w_0, w_1, \dots, w_n \}$) and $L$ refers to the set of generated labels after the encoding process. Then, the dependency graph $G$ will be converted into a well-formed sentiment graph $S$ to fully perform SSA. 

\subsection{Transition System}
\label{sec:tran}
A \textit{transition system} is defined by two components, namely \textit{state configurations} and \textit{actions} (also called \textit{transitions}). The transition-based framework aims to incrementally solve a problem by generating a sequence of actions  $A = {a_1, \dots, a_T}$. At each time step $t$, action $a_t$ will move the transition system from state configuration $s_{t-1}$ to a new one $s_{t}$, saving in the latter the partial output built so far. In our particular task, the transition system is designed for incrementally producing a dependency graph $G$ for an input sentence $X = w_0, w_1, \dots, w_n$ (where $w_0$ is the artificial root node).

\paragraph{State configurations} We follow  \citep{fernandez-gonzalez-gomez-rodriguez-2020-transition} and design a transition system that by means of two pointers parses the input text from left to right. More in detail, our transition system has state configurations of the form $s = \langle i, j, \Sigma \rangle$, where $i$ points at the word $w_i$ currently being processed, $j$ indicates the position of the last assigned head $w_j$ for $w_i$, and $\Sigma$ contains the set of already-built dependency arcs. The parsing process starts at the initial state configuration $s_{0} = \langle 1, -1,\emptyset \rangle$, where $i$ is pointing at the first input word $w_1$ in $X$ and, since no head has been assigned to $w_i$ yet, $j$ is set to $-1$ and $\Sigma$ is empty. Then, after applying a sequence of actions $A$, the transition system reaches a final configuration of the form $s_{T} = \langle n+1, -1, \Sigma \rangle$, where all the words have been processed 
and $\Sigma$ contains the edges of the dependency graph $G$ for the input sentence $X$.

\paragraph{Actions} The proposed transition system provides two transitions: 
\begin{itemize}
    \item An \textsc{Attach-to-}\textit{k} transition that attaches the current focus word $w_i$ to the head word at position $k$ (with $k \in [0,n]$ and $k \neq i$), generating a dependency arc from the head $w_k$ to the dependent $w_i$. 
    As a result, the transition system goes from state configurations $\langle i, j, \Sigma \rangle$ to $\langle i, k, \Sigma \cup \{w_k \rightarrow w_i\} \rangle$. This action is only allowed if the resulting dependency arc has not been created yet (i.e., $w_k \rightarrow w_i \notin \Sigma$) and $w_k$ is located to the right of the last assigned head $w_j$ for $w_i$ 
    (i.e., $j < k$). The latter condition is checked by storing the position of the last assigned head in pointer $j$ in the state configuration and it is necessary to keep the left-to-right order of the head assignment that was determined during training.
    Lastly, the created dependency arc is labeled by a jointly-trained classifier.
     \item A \textsc{Move} action that moves pointer $i$ one position to the right, pointing at the word $w_{i+1}$, and, since that unprocessed word has no assigned heads, $j$ is initialized to $-1$. This action is applied when the current focus word is completely processed and, as a result, moves from state configurations $\langle i, j, \Sigma \rangle$ to $\langle i+1, -1, \Sigma \rangle$.
\end{itemize} 
\noindent Please see in Table~\ref{tab:transitions} the required transition sequence (and resulting state configurations) for generating the dependency graph in Figure~\ref{fig_first}(b). The proposed transition system processes the input text from left to right by applying a sequence of \textsc{Move/Attach-to} actions that assign one or more heads to some tokens (leaving others unattached) and incrementally create in $\Sigma$ those dependency arcs that compose graph $G$. 

\begin{table}[!t]
\caption{Transition sequence 
for 
producing the dependency graph in Figure~\ref{fig_1}(b). Please note that \textsc{Att.-\textit{k}} stands for transition \textsc{Attach-to-}\textit{k}.} \label{tab:transitions}

\begin{tabular}{@{\hskip 1.0pt}l@{\hskip 5.0pt}c@{\hskip 5.0pt}l@{\hskip 1.0pt}c@{\hskip 1.0pt}c@{\hskip 1.0pt}c@{\hskip 1.0pt}}
\hline\noalign{\smallskip}
$t$ & $a_t$ & \ \ \ \ \ $s_t$ & $w_i$ & $w_j$  \\
\noalign{\smallskip}\hline\noalign{\smallskip}
 0 & - & $\langle 1, -1, \Sigma \rangle$ & Some$_1$ & - \\
1 & \textsc{Att.-}\textit{9} & $\langle 1, 9, \Sigma \cup \{  \text{Some}_1 \leftarrow \text{too}_9 \} \rangle$ & Some$_1$ & too$_9$  \\
2 & \textsc{Att.-}\textit{13} & $\langle 1, 13, \Sigma \cup \{ \text{Some}_1 \leftarrow \text{really}_{13} \} \rangle$ & Some$_1$ & really$_{13}$  \\
3 & \textsc{Move} & $\langle 2, -1, \Sigma \rangle$ & classmates$_2$ & -  \\
4 & \textsc{Att.-}\textit{1} & $\langle 2, 1, \Sigma \cup \{ \text{Some}_1\rightarrow\text{classmates}_{2} \} \rangle$ & classmates$_2$ & Some$_1$  \\
5 & \textsc{Move} & $\langle 3, -1, \Sigma \rangle$ & said$_3$ & -  \\
6 & \textsc{Move} & $\langle 4, -1, \Sigma \rangle$ & that$_4$ &  - \\
7 & \textsc{Move} & $\langle 5, -1, \Sigma \rangle$ & all$_5$ &  - \\
8 & \textsc{Att.-}\textit{9} & $\langle 5, 9, \Sigma \cup \{ \text{all}_5 \leftarrow \text{too}_{9} \} \rangle$ & all$_5$ & too$_9$  \\
9 & \textsc{Att.-}\textit{13} & $\langle 5, 13, \Sigma \cup \{ \text{all}_5 \leftarrow \text{really}_{13} \} \rangle$ & all$_5$ & really$_{13}$  \\
10 & \textsc{Move} & $\langle 6, -1, \Sigma \rangle$ & the$_6$ & -  \\
11 & \textsc{Att.-}\textit{5} & $\langle 6, 5, \Sigma \cup \{ \text{all}_5 \rightarrow \text{the}_6 \} \rangle$ & the$_6$ & all$_5$  \\
12 & \textsc{Move} & $\langle 7, -1, \Sigma \rangle$ & instructors$_7$ & -  \\
13 & \textsc{Att.-}\textit{5} & $\langle 7, 5, \Sigma \cup \{ \text{all}_5 \rightarrow \text{instructors}_7 \} \rangle$ & instructors$_7$ & all$_5$  \\
14 & \textsc{Move} & $\langle 8, -1, \Sigma \rangle$ & were$_8$ & -  \\
15 & \textsc{Move} & $\langle 9, -1, \Sigma \rangle$ & too$_9$ & -  \\
16 & \textsc{Att.-}\textit{0} & $\langle 9, 0, \Sigma \cup \{ \textsc{Root}_0 \rightarrow \text{too}_9 \} \rangle$ & too$_9$ & \textsc{Root}$_0$  \\
17 & \textsc{Move} & $\langle 10, -1, \Sigma \rangle$ & demanding$_{10}$ & -  \\
18 & \textsc{Att.-}\textit{9} & $\langle 10, 9, \Sigma \cup \{ \text{too}_9 \rightarrow \text{demanding}_{10} \} \rangle$ & demanding$_{10}$ & too$_9$  \\
19 & \textsc{Move} & $\langle 11, -1, \Sigma \rangle$ & \textbf{,}$_{11}$ & -  \\
20 & \textsc{Move} & $\langle 12, -1, \Sigma \rangle$ & but$_{12}$ & -  \\
21 & \textsc{Move} & $\langle 13, -1, \Sigma \rangle$ & really$_{13}$ & -  \\
22 & \textsc{Att.-}\textit{0} & $\langle 13, 0, \Sigma \cup \{ \textsc{Root}_0 \rightarrow \text{really}_{13} \} \rangle$ & really$_{13}$ & \textsc{Root}$_0$  \\
23 & \textsc{Move} & $\langle 14, -1, \Sigma \rangle$ & friendly$_{14}$ & -  \\
24 & \textsc{Att.-}\textit{13} & $\langle 14, 13, \Sigma \cup \{ \text{really}_{13} \rightarrow \text{friendly}_{14} \} \rangle$ & friendly$_{14}$ & really$_{13}$  \\
25 & \textsc{Move} & $\langle 15, -1, \Sigma \rangle$ & - & -  \\
\noalign{\smallskip}\hline
\end{tabular}

\end{table}

\subsection{Neural Architecture}
\label{sec:arch}
The aforementioned transition system relies on a Pointer Network \citep{Vinyals15} for selecting the action to be applied at each state configuration. This neural model employs an attention mechanism \citep{Bahdanau2015NeuralMT} to output, at each time step, a position from the input sentence. We will use that information to choose between the two available transitions. More in detail, the components of our neural architecture are described below:

\paragraph{Token representation} Similarly to the work by \cite{Barnes2021}, we exploit the concatenation of word ($e^{word}_i$), Part-of-Speech tag ($e^{pos}_i$), lemma ($e^{lemma}_i$) and character-level ($e^{char}_i$) embeddings for representing each token $w_i$ from the input sentence $X = w_1, \dots, w_n$ (without the root node $w_0$). In particular, we resort to \textit{convolutional neural networks} (CNNs)
\citep{ma-hovy-2016-end} for encoding characters inside $w_i$ and generate $e^{char}_i$. In addition, and following practically all previous studies in SSA, we further augment our model with deep contextualized word embeddings ($e^{\textsc{mBERT}}_i$) extracted from the multilingual variant of the pre-trained language model \textsc{BERT} (\textsc{mBERT})\citep{devlin-etal-2019-bert}. Concretely, we apply mean pooling (i.e., for each word, we use the average value of all available subword embeddings) to extract word-level representations from weights of 
the second-to-last layer. Therefore, the token representation $e_i$ is generated as follows:
 \begin{equation*}
e_i = e^{word}_i \oplus e^{pos}_i \oplus e^{lemma}_i \oplus e^{char}_i \oplus e^{\textsc{mBERT}}_i
 \end{equation*}
 \noindent where $\oplus$ denotes a concatenation operation.

\paragraph{Encoder} We exploit a three-layer bidirectional LSTM (BiLSTM) \citep{LSTM} to obtain a context-aware representation $c_i$ for each token vector $e_i$ from $E = e_1, \dots, e_n$: 
 \begin{equation*}
c_i  = \mathrm{BiLSTM}(e_i) = f_i \oplus b_i
\end{equation*}

\noindent where $f_i$ and $b_i$ are respectively the forward and backward hidden states of the last LSTM layer at the $i$th position. For representing the artificial root node $w_0$, we use a randomly-initialized  vector $c_0$. The final output of the encoding process is a sequence of encoder representations $C = c_0, c_1, \dots, c_n$.

\paragraph{Decoder} A unidirectional one-layer
LSTM equipped with an attention mechanism is deployed for implementing the transition-based decoding.
Given the transition system defined in Section~\ref{sec:tran}, each previous state configuration $s_{t-1}=\langle i, j, \Sigma \rangle$ is represented at time step $t$ as a decoder hidden vector $d_t$. This representation is obtained by feeding the LSTM with the 
element-wise summation of the respective encoder representations $c_i$ and $c_j$ of the focus word ($w_i$) and its last assigned head ($w_j$) if available:
\begin{align*}
r_t = c_{i} + c_{j}; d_t = \mathrm{LSTM}(r_t)
\end{align*}

\noindent where $r_t$ is the input vector of the decoder at time step $t$. Please note that $\Sigma$ is exclusively employed by the transition system to avoid creating duplicated edges and it is not used for the state configuration representation. It is also worth mentioning that, unlike in graph-based models, leveraging high-order information (such as the co-parent representation $c_j$ of a future head of $w_i$) does not harm runtime complexity in the transition-based framework. 

\begin{sidewaysfigure}[h]
\centering
\includegraphics[width=\textwidth]{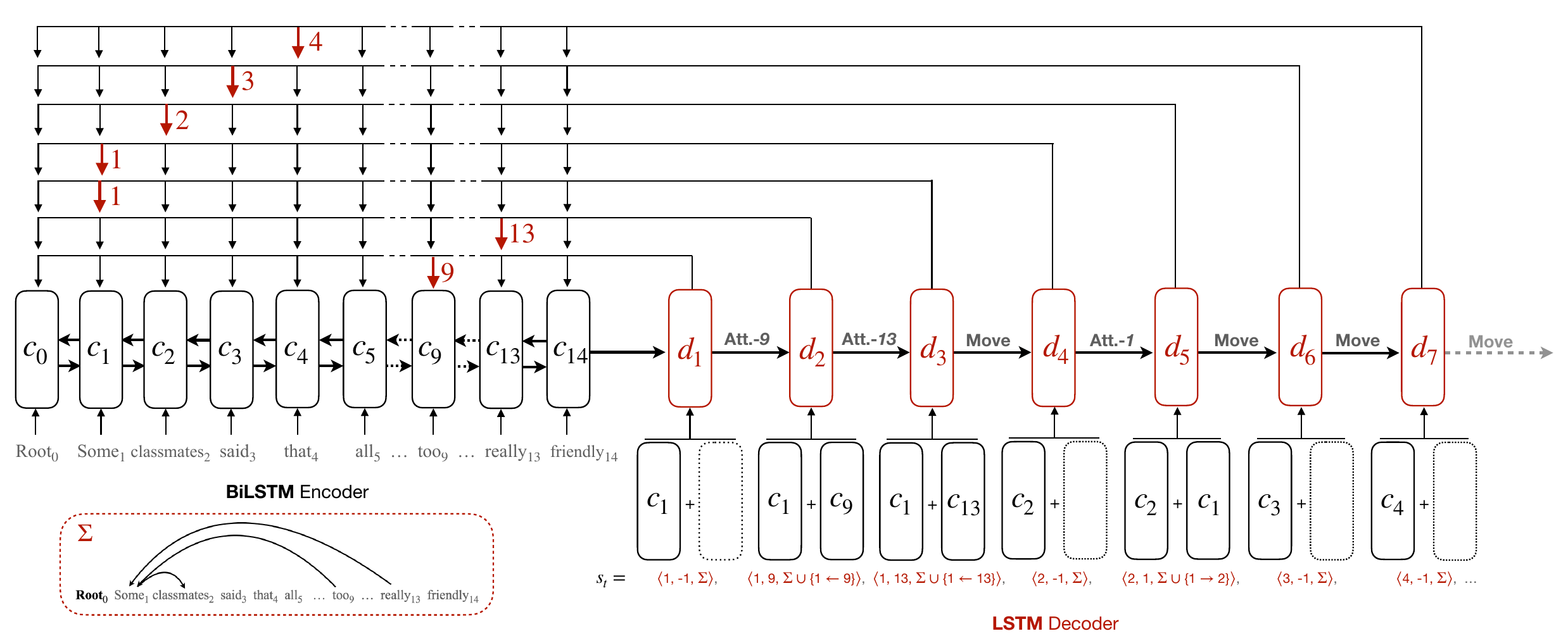}
\caption{The proposed neural architecture and decoding steps for partially producing the dependency graph in Figure~\ref{fig_first}. Please note that \texttt{Att.-\textit{k}} stands for transition \textsc{Attach-to-}\textit{k}.}
\label{fig_2}
\end{sidewaysfigure}

Subsequently, the attention mechanism is exploited for choosing the action $a_t$ to be applied to move the transition system from the current state configuration $s_{t-1}$ to the next one $s_{t}$. More in detail, this mechanism is implemented by a \textit{biaffine} scoring function \citep{DozatM17} that is used to compute the score between each input token $w_k$ (including the artificial root node $w_0$) and the state configuration representation $d_t$:
\begin{align*}
v^t_k = \mathrm{score}(d_t, c_k)= f_1(d_t)^T W f_2(c_k)\\
+U^Tf_1(d_t) + V^Tf_2(c_k) + b
\end{align*}
\noindent where each token $w_k$ is represented by the respective encoder vector $c_k$ (with $k \in [0,n]$), $W$ is the weight matrix of the bi-linear term, $U$ and $V$ are the weight tensors of the linear terms, $b$ is the bias vector and 
$f_1(\cdot)$ and $f_2(\cdot)$ are two one-layer perceptrons with ELU activation \citep{ELU} to obtain lower-dimensionality and avoid overfitting. Vector $v^t$ of length $n+1$ is then normalized to output in the attention vector $\alpha^t$ a \textsc{softmax} distribution with dictionary size equal to $n+1$ :
\begin{equation*}
\alpha^t_k = \frac{\text{exp}(v^t_k)}{\sum _{k=0}^{n} \text{exp}(v^t_k)} 
\end{equation*}
\noindent Lastly, attention scoring vector $\alpha^t$ is used for choosing the highest-scoring position $k^*$ from the input:
\begin{equation*}
k^* = \mathrm{Argmax}_{0\leq k \leq n}(\alpha^t_0, \dots, \alpha^t_n)
\end{equation*}
The pointed position $k^*$ is then utilized by the transition system to select the action $a_t$ to be applied on the previous state configuration $s_{t-1}=\langle i, j, \Sigma \rangle$:
\begin{itemize}
    \item if $k^*=i$, it means that $w_i$ is completely processed and the transition system has to read the next word from the input. Therefore, $a_t$ must be a \textsc{Move} transition, shifting the focus word pointer $i$ to the next word and setting $j$ to -1.
    \item On the contrary, if $k^* \neq i$, then an \textsc{Attach-to} transition parameterized with $k^*$ is considered, creating a dependency arc from the head token in position $k^*$ ($w_{k^*}$) to the current focus word $w_i$. 
\end{itemize}
Please note that, if those conditions required by the \textsc{Attach-to} action are not satisfied, the next highest-scoring position in $\alpha^t$ will be considered for choosing again between the two available transitions. In Figure~\ref{fig_2}, we depict a sketch of the proposed neural architecture and decoding steps for partially producing the dependency graph in Figure~\ref{fig_first}(b).

\paragraph{Dependency Labeler}
Each time the transition system attaches the current focus word $w_i$ to a head, a multi-class classifier is applied for labeling the resulting dependency arc. The aforementioned scoring function is exploited again for implementing the labeler. In particular, for each  available dependency label $l \in L$, we
compute the score of assigning $l$ to the predicted dependency arc between $w_i$ (encoded as $d_t$) and the head token $w_{k^*}$ (represented by $c_{k^*}$):
\begin{align*}
u^{t}_l = \mathrm{score}(d_t, c_{k^*}, l)= g_1(d_t)^T W_l g_2(c_{k^*})\\
+U_l^Tg_1(d_t) + V_l^Tg_2(c_{k^*}) + b_l
\end{align*}
\noindent where $W_l$ is a weight matrix, $U_l$ and $V_l$ are weight tensors and $b_l$ is the bias vector exclusively used for each label $l$, and $g_1(\cdot)$ and $g_2(\cdot)$ are two one-layer perceptrons with ELU activation.

Being $m$ the number of labels in $L$, we normalize vector $u^t$ to output in vector $\beta^t$ a \textsc{softmax} distribution with dictionary size equal to $m$:
\begin{equation*}
\beta^t_l = \frac{\text{exp}(u^t_l)}{\sum _{l=1}^{m} \text{exp}(u^t_l)} 
\end{equation*}
\noindent Finally, we choose the highest-scoring label $l^*$ from vector $\beta^t$ for tagging the resulting dependency arc:
\begin{equation*}
l^* = \mathrm{Argmax}_{1\leq l \leq m}(\beta^t_1, \dots, \beta^t_m)
\end{equation*}

\paragraph{Training objectives}
The proposed transition-based model aims to predict the labeled dependency graph $G$ 
for the input sentence $X$. This 
can be decomposed into two different objectives: transition sequence prediction and label prediction.

First, the transition system is trained to optimize the probability of producing unlabeled dependency graphs $G'$ given sentences: $P_\theta (G'|X)$. This probability can be factorized by transforming
the dependency graph $G'$ into a sequence of transitions $A = a_1, \dots, a_T$, which (in the proposed neural architecture) is equivalent to the
sequence of indices $k_1, \dots, k_T$ required to output $A$. In addition, since the Pointer Network is an auto-regressive model, each prediction $k_t$ at time step $t$ is conditioned by previous decisions ($k_{<t}$). The resulting probability remains as follows:
\begin{equation*}
\label{prob}
    P_\theta(G'|X) = \prod_{t=1}^{T} P_\theta (k_t | k_{<t},X)
\end{equation*}

Based on this probability, we define the first objective as a cross-entropy loss 
that minimizes the negative log-likelihood of predicting the correct sequence of indices $k_1, \dots, k_T$:
\begin{equation*}
    \mathcal{L}_{tran}(\theta) = - \sum_{t=1}^{T} \text{log} P_\theta (k_t | k_{<t},X)
\end{equation*}

On the other hand, the second objective consists of labeling each predicted arc from the unlabeled dependency graph $G'$ with the gold label $l$. Thus, the
labeler is trained 
to minimize the negative log-likelihood of assigning the correct label $l$ for each predicted arc from head $w_{k_t}$ to dependent $w_i$.
The cross-entropy loss is then computed as follows:
\begin{equation*}
\mathcal{L}_{label}(\theta) = - \sum_{t=1}^{T} \text{log} P_\theta (l | w_{k_t},w_i)
\end{equation*}

Lastly, 
we jointly train the two components of our model by optimizing the sum of their objectives:
\begin{equation*}
\mathcal{L}(\theta) = \mathcal{L}_{tran}(\theta) + \mathcal{L}_{label}(\theta)
\end{equation*}
where $\theta$ stands for model parameters.

\section{Experiments}
\label{sec:experiments}
We conduct an extensive evaluation of the proposed transition-based approach on standard SSA benchmarks and undertake a thorough comparison against other dependency-based methods and task-specific approaches that can directly build sentiment graphs. In this section, we first describe the datasets, training setup and evaluation metrics and then report the main results. We additionally include an in-depth analysis of the impact of the different components of our model, the effect of using syntactic information in the encoding scheme and the efficiency of our transition-based model. 

\subsection{Datasets}
We follow the  original work by \cite{Barnes2021} and conduct experiments on five SSA datasets, covering four different languages.  More in detail, NoReC$_{\text{Fine}}$ \citep{ovrelid-etal-2020-fine} is the largest dataset and consists of multi-domain professional reviews in Norwegian; MultiB$_{\text{EU}}$ and MultiB$_{\text{CA}}$ \citep{barnes-etal-2018-multibooked} collect hotel reviews in Basque and Catalan, respectively; MPQA \citep{Wiebe2005AnnotatingEO} is compounded by news wire text in English; and DS$_{\text{Unis}}$ \citep{toprak-etal-2010-sentence} is derived from reviews of online universities and e-commerce in English (although only the former are included). 
Moreover, we report results on two different versions of the MPQA dataset in order to compare with prior related work. In particular, we include the initial version used by \cite{Barnes2021} and the extension proposed by \cite{Samuel2022} (which we name as MPQA$_\text{v2}$). 
We use the preprocessing scripts publicly available with default training, development and test splits.\footnote{\url{github.com/jerbarnes/sentiment_graphs/tree/master/data/sent_graphs}}
In Table~\ref{tab:stats}, we summarize statistics for each dataset.
We can see as NoReC$_{\text{Fine}}$, MPQA and MPQA$_\text{v2}$ are not only significantly larger with respect to the other datasets but also include more complex sentiment graphs: NoReC$_{\text{Fine}}$ annotates the largest  proportion of opinion expressions and targets, and MPQA and MPQA$_\text{v2}$ provide a notable amount of opinion holders in comparison to the other benchmarks. It is also worth mentioning that, while NoReC$_{\text{Fine}}$, MultiB$_{\text{EU}}$ and MultiB$_{\text{CA}}$ only include positive and negative polarities; DS$_{\text{Unis}}$, MPQA and MPQA$_\text{v2}$ additionally annotate a neutral polarity.


\begin{table}[!t]
\caption{Data statistics for each benchmark. For each split, we report the number of sentences, holders, targets and expressions, as well as the distribution of polarity (positive, neutral or negative).}
\label{tab:stats}
\centering
\begin{tabular}{@{\hskip 2pt}l@{\hskip 5pt}l@{\hskip 10pt}r@{\hskip 10pt}r@{\hskip 10pt}r@{\hskip 10pt}r@{\hskip 10pt}r@{\hskip 2pt}}
\hline
Dataset & Split & Sent. & Holders & Targets & Expr. & Polarity (+/neu/-) \\
\hline
NoReC$_{\text{Fine}}$ & train & 8,634 & 898 & 6,778 & 8,448 &  5,684/ 0 / 2,756\\
& dev & 1,531 & 120 & 1,152 & 1,432 &  988 / 0 / 443\\
& test & 1,272 & 110 & 993 & 1,235 &  875 / 0 / 358\\
MultiB$_{\text{EU}}$ & train & 1,064 & 205 & 1,285 & 1,684 & 1,406 / 0 / 278 \\
& dev & 152 & 33 & 153 & 204 & 168 / 0 / 36 \\
& test & 305 & 58 & 337 & 440 & 375 / 0 / 65 \\
MultiB$_{\text{CA}}$ & train & 1,174 & 169 & 1,695 & 1,981 & 1,272 / 0 / 708 \\
& dev & 168 & 15 & 211 & 258 & 151 / 0 / 107 \\
& test & 336 & 52 & 430 & 518 & 313 / 0 / 204 \\
MPQA & train & 4,500 & 1,306 & 1,382 & 1,656 & 675 / 271 / 658 \\
& dev & 1,622 & 377 & 449 & 552 & 241 / 105 / 202 \\
& test & 1,681 & 371 & 405 & 479 & 166 / 89 / 199 \\
MPQA$_\text{v2}$ & train & 5,873 & 1,431 & 1,487 & 1,715 & 671 / 337 / 698 \\
& dev & 2,063 & 414 & 503 & 581 & 223 / 126 / 216 \\
& test & 2,112 & 434 & 462 & 518 & 159 / 82 / 223 \\
DS$_{\text{Unis}}$ & train & 2,253 & 65 & 836 & 836 & 349 / 104 / 383 \\
& dev & 232 & 9 & 104 & 104 & 31 / 16 / 57 \\
& test & 318 & 12 & 142 & 142 & 59 / 12 / 71 \\
\hline
\end{tabular}
\end{table}


\subsection{Evaluation Metrics}
We adopt the evaluation metrics proposed by \cite{Barnes2021} and directly use their scoring script.\footnote{\url{github.com/jerbarnes/sentiment_graphs/blob/master/src/F1_scorer.py}} Specifically, we use the following metrics:
\begin{itemize}
    \item \textit{Token-level F$_1$ scores for holder, target and expression spans (Span F$_1$).} These metrics evaluate the performance on span identification for each opinion component at the token level.
    \item \textit{Targeted F$_1$ score.} This metric is widely used in aspect-based sentiment analysis and it measures the exact detection of the opinion target together with the correct polarity. 
    \item \textit{Labeled and unlabeled F$_1$ scores in dependency graph parsing (LF$_1$, UF$_1$).} These graph-level metrics evaluate the performance in predicting correct arcs with and without the correct label, respectively. They are only intended for dependency-based methods and are used to measure the accuracy in producing the intermediate dependency graph, before recovering the final sentiment graph.
    \item \textit{Sentiment graph F$_1$ score (SF$_1$) and Non-polar sentiment graph F$_1$ score (NSF$_1$).} These metrics provide the overall performance in SSA. In particular, NSF$_1$ considers that opinion tuples are only compounded by the holder, target and expression, while SF$_1$ additionally includes the opinion polarity. In both metrics, a true positive is an exact match at the graph level, weighting the overlap in predicted and gold spans for each opinion component, and averaged across all three (holder, target and expression) spans.
\end{itemize}

\subsection{Training Details}
For a fair comparison with earlier works, we initialize word and lemma embeddings with word2vec skip-gram vectors \citep{fares-etal-2017-word} (with dimensions 300 for English and 100 for the remaining languages). We also randomly initialize 100-dimensional character-level embeddings and use 100 filters with a window size of 3 plus max-pooling for CNNs. For all languages, we additionally leverage 768-dimensional token-level representations extracted  from the multilingual pre-trained language model \textsc{mBERT} \citep{devlin-etal-2019-bert} (concretely, {\tt bert-base-multilingual-cased}). As in most of the previous studies in SSA, we follow a greener and less resource-consuming strategy and do not fine-tune BERT-based embeddings during training.

For all datasets, we use most hyper-parameters from \citep{fernandez-gonzalez-gomez-rodriguez-2020-transition} without further exploration. Specifically, we optimize our neural architecture with Adam 
and initial learning rate of $\eta_0 =$ 0.001, $\beta_1 =$ 0.9 and $\beta_2=$ 0.9. For reducing the impact of the gradient exploding effect, we apply a fixed decay rate of 0.75 and a gradient clipping of 5.0. 
In addition, we exploit LSTMs with 512-dimensional hidden states for both the encoder and decoder. We apply a 0.33 dropout to all embeddings, and 
between hidden states and layers. 
All models are trained for 600 epochs with batch size 32, and beam-search decoding with beam size 5 is exploited for all experiments. Similarly to \citep{Barnes2021}, we choose the checkpoint with the highest LF$_1$ score on the development set and execute 5 times all models with different random initializations, reporting the average and standard deviation (the latter is presented in Appendix \ref{appendix}).


\subsection{Comparing Methods}
We first compare our approach to other existing dependency-based methods, namely graph-based models by \cite{Barnes2021} and \cite{Peng2021}. For undertaking a fair comparison to these baselines, we resort to the same dependency-based representations, and employ a similar BiLSTM encoder and token representation (including frozen embeddings extracted from \textsc{mBERT}). Finally, while the accuracy obtained by the head-first and head-final encoding schemes is available for the model by \cite{Barnes2021}; \cite{Peng2021} only report, for each benchmark, the performance of the encoding strategy that obtains the highest scores.

In order to put into context our transition-based method, we additionally compare it against recent task-specific approaches that achieve the best accuracy to date in SSA. Among them, we include models developed by \cite{Shi2022}, \cite{Zhai2023}, \cite{Zhou2024} and \cite{Samuel2022}, as well as the transition-based approach designed by \cite{Xu2022}. All these methods follow the common practice of incorporating frozen word-level embeddings from \textsc{mBERT}, except for the system by \cite{Samuel2022}. This approach exploits deep contextualized representations from XLM-R \citep{conneau-etal-2020-unsupervised} and specifically updates them for the SSA task. For a fairer comparison, we also report the accuracy obtained by this model without the expensive fine-tuning.

\subsection{Results and Discussion}


In Table~\ref{tab:results}, we present the comparison of our transition-based model with prior dependency-based methods. Our approach outperforms the original graph-based model by \cite{Barnes2021} across all benchmarks, not only in full SSA (indicated by the SF$_1$ metric) but also in other span identification metrics and, especially, in target extraction plus polarity classification (measured by the Targeted F$_1$ score). Regarding the state-of-the-art model by \cite{Peng2021}, it is surpassed by the proposed transition-based method in four out of five datasets in terms of SF$_1$ and in all benchmarks according to the Targeted F$_1$ score. It is also worth mentioning that our approach 
obtains notable accuracy gains in SSA on the largest and more complex benchmarks (NoReC$_\text{Fine}$ and MPQA). On the other hand, it can be seen as the graph-based model by \cite{Peng2021} achieves higher scores than our method in span extraction (especially on NoReC$_\text{Fine}$ and DS$_\text{Unis}$) and also excels in dependency graph metrics in practically all datasets. However, as noticed by \cite{Barnes2021}, dependency and sentiment graph metrics do not correlate, meaning that achieving the best accuracy on the intermediate dependency-based representation does not necessarily lead to the best SF$_1$ score on the recovered sentiment graph. On the contrary, we do see a correlation between the Targeted F$_1$ metric and the SF$_1$ score: while having lower scores on span identification and dependency graph parsing, our approach excels in identifying the opinion target and assigning the correct sentiment polarity, which seem to
have a crucial impact on the full SSA task. This can be probably explained by the fact that performance on polarity classification (a critical sub-task in full SSA) is exclusively measured by the Targeted F$_1$ and SF$_1$ metrics.

Regarding the performance of the two encoding methods explored in our approach, we can observe in Table~\ref{tab:results} as the head-first dependency-based representation leads to a higher accuracy across practically all datasets and metrics, except for 
the MultiB$_{\text{EU}}$ benchmark. 
On this dataset, the head-first method obtains better scores in holder and expression identification as well as dependency graph parsing, but it is surpassed by the head-final encoding in the most important sentiment metrics (Targeted F$_1$ and SF$_1$). Finally, it can be also noticed as, in some datasets, the encoding strategy that yields the best scores in graph-based models differs from the top-performing representation in our transition-based method. This behavior is expected due to the fact that each approach builds the intermediate dependency graph in a different manner. \new{In fact, we hypothesize that the superior performance of the head-first encoding over the head-final strategy in our approach stems from the left-to-right nature of our transition system: in head-first structures, the head word precedes its dependents and is therefore already available to the parser when establishing dependency links.}





\begin{sidewaystable}[!t]
\caption{Accuracy comparison of our work with graph-based baselines for approaching SSA as dependency \added{graph} parsing.}
\label{tab:results}
\centering
\begin{tabular}{lllcccccccc}
\hline
 Dataset   &   Model & Encoding         &   \multicolumn{3}{c}{Span F$_1$} &   Targeted &   \multicolumn{2}{c}{Depen. Graph} &   \multicolumn{2}{c}{Sentim. Graph} \\
  &    &    &   Holder &   Target &   Exp. &  F$_1$ &   UF$_1$ &   LF$_1$ &   NSF$_1$ &   SF$_1$ \\
\hline
\textbf{NoReC$_\text{Fine}$}   & \citep{Barnes2021} & head-first &     51.1 &     50.1 &  54.4 &  30.5 & 39.2 & 31.5 &  37.0 &  29.5 \\
& \citep{Barnes2021} & head-final &     60.4 &     54.8 &  55.5 &  31.9 & \textbf{48.0} & 37.7 &  39.2 &  31.2 \\
& \citep{Peng2021} & head-final &     \textbf{63.6} &     \textbf{55.3} &  \textbf{56.1} &  31.5 & - & \textbf{40.4} &  - &  31.9 \\
& \textbf{This work} & head-first &    55.8 & 53.8 & 52.0 & \textbf{41.4} & 39.9 & 32.8 & 43.6 & \textbf{35.8} \\
& \textbf{This work} & head-final &   57.5 & 54.3 & 51.5 & 36.1 & 45.0 & 36.5 & \textbf{44.2} & \textbf{35.8} \\

\hline
\textbf{MultiB$_\text{EU}$}   & \citep{Barnes2021} & head-first &     60.4 &  64.0 &  73.9 &  57.8 & 64.6 & 60.0 &  58.0 &  54.7 \\
& \citep{Barnes2021} & head-final &     60.5 &  64.0 &  72.1 &  56.9 & 60.8 & 56.0 &  58.0 &  54.7 \\
& \citep{Peng2021} & head-first &     65.8 &     \textbf{71.0} &  76.7 &  59.6 & - & \textbf{66.1} &  - &  \textbf{62.7} \\
& \textbf{This work} & head-first &    \textbf{67.0} & 67.7 & \textbf{77.2} & 59.1 & \textbf{68.2} & 63.4 & 63.7 & 60.2 \\
& \textbf{This work} & head-final &    66.6 & 67.9 & 76.6 & \textbf{60.5} & 66.9 & 61.4 & \textbf{63.8} & 60.4 \\
\hline
\textbf{MultiB$_\text{CA}$}   & \citep{Barnes2021} & head-first &     43.0 &  72.5 &  71.1 &  55.0 & 66.8 & 62.1 &  62.0 &  56.8 \\
& \citep{Barnes2021} & head-final &     37.1 &  71.2 &  67.1 &  53.9 & 62.7 & 58.1 &  59.7 &  53.7 \\
& \citep{Peng2021} & head-first &     46.2 &     74.2 &  71.0 &  60.9 & - & \textbf{64.5} &  - &  59.3 \\
& \textbf{This work} & head-first &    54.1 & \textbf{75.4} & \textbf{71.4} & \textbf{64.0} & \textbf{68.5} & 64.4 & \textbf{67.0} & \textbf{61.0}   \\
& \textbf{This work} & head-final &     \textbf{55.7} & 73.9 & 69.6 & 57.0 & 65.2 & 60.2 & 64.7 & 59.0 \\
\hline
\textbf{MPQA}   & \citep{Barnes2021} & head-first &     43.8 &  \textbf{51.0} &  \textbf{48.1} &  33.5 & 40.0 & 36.9 &  24.5 &  17.4 \\
& \citep{Barnes2021} & head-final &     46.3 &  49.5 &  46.0 &  18.6 & \textbf{41.4} & 38.0 &  26.1 &  18.8 \\
& \citep{Peng2021} & head-first &     \textbf{47.9} &     50.7 &  47.8 &  33.7 & - & \textbf{38.6} &  - &  19.1 \\
& \textbf{This work} & head-first &  46.2 & 49.7 & 44.7 & \textbf{33.9} & 40.5 & 37.6 & \textbf{28.7} & \textbf{22.2} \\
& \textbf{This work} & head-final &    43.8 & 46.4 & 45.2 & 18.1 & 38.0 & 35.4 & 25.9 & 19.7 \\


\hline
\textbf{DS$_\text{Unis}$}   & \citep{Barnes2021} & head-first &     28.0 &  39.9 &  40.3 &  26.7 & 35.3 & 31.4 &  31.0 &  25.0 \\
& \citep{Barnes2021} & head-final &      37.4 &  42.1 &  \textbf{45.5} &  29.6 & \textbf{38.1} & 33.9 &  34.3 &  26.5 \\
& \citep{Peng2021} & head-final &     \textbf{50.0} &     \textbf{44.8} &  43.7 &  30.7 & - & \textbf{35.0} &  - &  27.4 \\
& \textbf{This work} & head-first &   40.4 & 44.3 & 44.3 & \textbf{32.5} & 37.3 & 32.5 & \textbf{35.9} & \textbf{29.9} \\ 
& \textbf{This work} & head-final &  43.2 & 44.4 & 43.6 & 28.7 & 36.3 & 30.3 & 32.7 & 25.6 \\
\hline
\end{tabular}
\end{sidewaystable}

We additionally compare our proposal against top-performing SSA methods in Table~\ref{tab:sota}, including 
task-specific approaches. While relying on a general-purpose parser, our model ranks third in terms of SF$_1$ on MultiB$_\text{EU}$ and MultiB$_\text{CA}$ datasets and achieves the best SF$_1$ score to date on the MPQA benchmark. Among those models that report Targeted F$_1$ score, the proposed system also obtains the highest score on four of out six datasets. Regarding span identification, our model yields a competitive accuracy and even outperforms all task-specific approaches in identifying holder nodes on MultiB$_\text{EU}$ and MultiB$_\text{CA}$ datasets. Finally, it is worth noting that, while notably simpler, our approach improves over the transition-based model by \cite{Xu2022} in five out of six datasets according to the SF$_1$ metric; and the model by \cite{Samuel2022} (enhanced with fine-tuned \textsc{XLM-R}-based embeddings) is surpassed by our algorithm in SSA without polarity identification (indicated by the NSF$_1$ score) on smaller datasets (MultiB$_\text{EU}$, MultiB$_\text{CA}$ and DS$_\text{Unis}$).


\added{Lastly, while not fine-tuning hyper-parameters for individual datasets, our approach delivers a strong and consistent performance in SSA across benchmarks, which suggests that
the proposed transition-based method is robust and can be directly applied to other datasets.}

\begin{table}[h]
\caption{Accuracy comparison of our transition-based approach against 
state-of-the-art methods. Those models
marked with $^*$ are enhanced with frozen XLM-R-based representations and those marked with $^\dagger$ exploit fine-tuned XLM-R-based embeddings. For our approach and \citep{Barnes2021}, we only include the variant with the encoding strategy that achieves the best SF$_1$ score.} \label{tab:sota}
\centering
\begin{tabular}{@{\hskip 1.0pt}l@{\hskip 2.0pt}l@{\hskip 2.0pt}c@{\hskip 5.0pt}c@{\hskip 5.0pt}c@{\hskip 8.0pt}c@{\hskip 8.0pt}c@{\hskip 5.0pt}c@{\hskip 1.0pt}}
\hline
 Dataset   &   Model         &   \multicolumn{3}{c}{Span F$_1$} &   Target. &     \multicolumn{2}{c}{Sent. G.} \\
  &        &   Hold. &   Targ. &   Exp. &  F$_1$ &   NSF$_1$ &   SF$_1$ \\
\hline
\textbf{NoReC$_\text{Fine}$} & \citep{Barnes2021} &     60.4 &     54.8 &  55.5 &  31.9 & 39.2 &  31.2 \\
& \citep{Peng2021} &     63.6 &     55.3 &  56.1 &  31.5 &  - &  31.9 \\
& \citep{Shi2022} &     60.9 &     53.2 &  61.0 &  38.1 &  46.4 &  37.6 \\
& \citep{Xu2022} &     61.7 &     56.3 &  60.4 &  37.4 &  \textbf{49.7} &  37.8 \\
& \citep{Samuel2022}$^*$ &     48.3 &     51.9 &  57.9 &  - &  41.8 &  35.7 \\
& \citep{Samuel2022}$^\dagger$ &     65.1 &     \textbf{58.3} &  60.7 &  - &  47.8 &  \textbf{41.6} \\
& \citep{Zhai2023} & 66.3 & 54.3 & 61.4 & - & 47.7 & 39.6 \\
& \citep{Zhou2024} & \textbf{67.4} & 54.5 & \textbf{62.7} & - & 49.5 & 41.5 \\
& \textbf{This work} &   55.8 & 53.8 & 52.0 & \textbf{41.4} & 43.6 & 35.8 \\

\hline
\textbf{MultiB$_\text{EU}$}   & \citep{Barnes2021}  &     60.4 &  64.0 &  73.9 &  57.8 &  58.0 &  54.7 \\
& \citep{Peng2021} &     65.8 &     \textbf{71.0} &  \textbf{76.7} &  59.6 &  - &  \textbf{62.7} \\
& \citep{Shi2022} &     62.8 &  65.6 &  75.2 &  60.9 &  61.1 &  58.9 \\
& \citep{Xu2022}  &     62.4 &   66.8 & 75.5 &  \textbf{61.5} &  62.7 &  56.8 \\
& \citep{Samuel2022}$^*$ &     55.5 &     58.5 &  68.8 &  - &  53.1 &  51.3 \\
& \citep{Samuel2022}$^\dagger$ &     64.2 &     67.4 &  73.2 &  - &  62.5 &  61.3 \\
& \citep{Zhai2023} & 63.4 & 66.9 & 75.4 & - & 63.5 & 60.4 \\
& \citep{Zhou2024} & 65.5 & 68.2 & 75.8 & - & \textbf{65.7} & \textbf{62.7} \\
& \textbf{This work} &      \textbf{66.6} & 67.9 & 76.6 & 60.5 & 63.8 & 60.4 \\

\hline
\textbf{MultiB$_\text{CA}$}   & \citep{Barnes2021} &     43.0 &  72.5 &  71.1 &  55.0 &  62.0 &  56.8 \\
& \citep{Peng2021} &     46.2 &     74.2 &  71.0 &  60.9 &  - &  59.3 \\
& \citep{Shi2022} &     47.4 &  73.8 &  71.8 &  60.6 &  64.2 &  59.8 \\
& \citep{Xu2022}  &     46.8 & 74.5 & 72.4 &  60.0 &  66.8 &  59.2 \\
& \citep{Samuel2022}$^*$ &     39.8 &     69.2 &  66.3 &  - &  60.2 &  57.6 \\
& \citep{Samuel2022}$^\dagger$ &     48.0 &     72.5 &  68.9 &  - &  65.7 &  \textbf{63.3} \\
& \citep{Zhai2023} & 47.5 & 74.2 & 72.2 & - & 67.4 & 61.0 \\
& \citep{Zhou2024} & 50.3 & 75.2 & \textbf{74.7} & - & \textbf{69.7} & 62.8 \\
& \textbf{This work} &    \textbf{54.1} & \textbf{75.4} & 71.4 & \textbf{64.0} & 67.0 & 61.0   \\

\hline
\textbf{MPQA}   & \citep{Barnes2021} &     43.8 &  51.0 &  48.1 &  33.5 &  24.5 &  17.4 \\
& \citep{Peng2021} &     47.9 &     50.7 &  47.8 & 33.7 &  - &  19.1 \\
& \citep{Shi2022} &     44.1 &  51.7 &  47.8 &  23.3 &  28.2 &  21.6 \\
& \citep{Xu2022}  &     \textbf{48.3} &  \textbf{52.8} & \textbf{50.0} &  \textbf{36.5} &  \textbf{28.9} &  20.1 \\
& \textbf{This work} &  46.2 & 49.7 & 44.7 & 33.9 & 28.7 & \textbf{22.2} \\
\hline
\textbf{MPQA$_\text{v2}$}   & \citep{Samuel2022}$^*$ &     44.0 &     49.0 &  46.6 &  - &  30.7 &  23.1 \\
& \citep{Samuel2022}$^\dagger$ &     \textbf{55.7} &     \textbf{64.0} &  \textbf{53.5} &  - &  \textbf{45.1} &  \textbf{34.1} \\
& \citep{Zhai2023} & 47.3 & 58.9 & 48.0 & - & 36.8 & 30.5 \\
& \citep{Zhou2024} & 51.2 & 60.2 & 48.2 & - & 40.1 & 32.4 \\
& \textbf{This work} &   53.1 & 57.3 & 48.5 & \textbf{41.1} & 34.6 & 27.0   \\

\hline
\textbf{DS$_\text{Unis}$}   & \citep{Barnes2021} &      37.4 &  42.1 &  45.5 &  29.6 &  34.3 &  26.5 \\
& \citep{Peng2021} &     \textbf{50.0} &     44.8 &  43.7 &  30.7 &  - &  27.4 \\
& \citep{Shi2022} &     43.7 &  49.0 &  42.6 &  31.6 &  36.1 &  31.1 \\
& \citep{Xu2022}  &     42.8 &   48.3 & 47.8 &  31.5 &  38.4 &  29.3 \\
& \citep{Samuel2022}$^*$ &     13.8 &     37.3 &  33.2 &  - &  24.5 &  21.3 \\
& \citep{Samuel2022}$^\dagger$ &     42.2 &     40.6 &  39.3 &  - &  33.2 &  \textbf{31.2} \\
& \citep{Zhai2023} & 44.2 & 50.2 & 46.6 & - & 38.0 & 33.2 \\
& \citep{Zhou2024} & 44.4 & \textbf{51.0} & \textbf{48.2} & - & \textbf{40.1} & 35.7 \\
& \textbf{This work} &       40.4 & 44.3 & 44.3 & \textbf{32.5} & 35.9 & 29.9 \\ 

\hline
\end{tabular}
\end{table}

\subsection{Ablation Study}
In order to better understand the contribution of each component, we conduct an ablation study of our neural architecture. Specifically, we analyze the influence of  
beam-search decoding, co-parent features (obtained by concatenating the encoder representations of the previously-assigned head   $c_j$ to the current focus word $c_i$), Part-of-Speech (PoS) tag embeddings, lemma embeddings, character-level embeddings and \textsc{mBERT}-based representations. 
We just 
perform the ablation study on development splits of MultiB$_{\text{EU}}$ and DS$_{\text{Unis}}$ datasets with the head-first graph encoding and report the results in Table~\ref{tab:study}.

\begin{table}[!t]
\caption{Ablation study on the development split with the head-first graph encoding. Apart from indicating the best score with bold numbers for each metric, we use italic numbers to mark those results that, while not being the highest score, surpass the full model accuracy.} 
\label{tab:study}
\centering
\begin{tabular}{@{\hskip 1.0pt}l@{\hskip 5.0pt}c@{\hskip 5.0pt}c@{\hskip 5pt}c@{\hskip 8pt}c@{\hskip 8pt}c@{\hskip 5pt}c@{\hskip 8.0pt}c@{\hskip 5.0pt}c@{\hskip 1.0pt}}
\hline
   Model        &   \multicolumn{3}{c}{Span F$_1$} &   Targ. &   \multicolumn{2}{c}{Dep. G.}  &   \multicolumn{2}{c}{Sent. G.} \\
      &   Hold. &   Targ. &   Exp. &  F$_1$ &   UF$_1$ &   LF$_1$ &   NSF$_1$ &   SF$_1$ \\
\hline
 \textbf{MultiB$_\text{EU}$}  full &    53.6 & 72.1 & 71.1 & \textbf{64.3} & 64.6 & 59.2 & 65.8 & 61.1 \\
\ \  w/o PoS emb. & 52.0 & 70.1 & \textit{71.5} & 62.9 & 64.6 & 59.2 & 63.6 & 59.0 \\
 \ \  w/o char. emb. & 52.5 & 70.9 & 71.0 & 64.2 & 64.5 & \textit{59.3} & 63.4 & 59.3 \\
 \ \  w/o lemma emb. & 50.8 & 70.6 & \textit{71.6} & 62.8 & 64.4 & 58.5 & 64.0 & 59.5 \\
 \ \  w/o co-parent & 52.7 & 71.0 & \textbf{72.5} & 63.4 & \textit{64.7} & \textit{59.6} & 64.0 & 59.9 \\
\ \  w/o beam search & \textbf{54.4} & 70.2 & \textit{71.9} & 62.8 & \textbf{65.2} & \textit{59.9} & 64.8 & 60.4 \\
 \ \  w/o \textsc{mBERT} emb. & 48.1 & \textbf{74.7} & \textit{72.1} & 63.9 & \textbf{65.2} & \textbf{60.9} & \textbf{67.4} & \textbf{62.4} \\
\hline
 \textbf{DS$_\text{Unis}$} full &   20.0 & 48.7 & \textbf{48.7} & 29.3 & \textbf{46.4} & \textbf{38.8} & 35.9 & \textbf{24.4} \\
 \ \  w/o PoS emb. & 17.3 & 47.4 & 47.1 & \textit{29.9} & 43.4 & 35.1 & 35.4 & 24.0 \\
 \ \  w/o char. emb. & \textbf{20.6} & 48.4 & \textbf{48.7} & 29.2 & 45.2 & 37.0 & \textbf{36.1} & 24.1 \\
 \ \  w/o lemma emb. & 17.0 & 48.3 & 46.3 & 27.1 & 43.0 & 33.9 & 34.1 & 20.5 \\
 \ \  w/o co-parent & 8.0 & 47.8 & 47.7 & 28.2 & 44.6 & 35.1 & 34.9 & 23.1 \\ 
 \ \  w/o beam search & 7.1 & \textbf{49.1} & 47.6 & \textbf{32.1} & 44.2 & 35.8 & 35.7 & 23.5 \\
 \ \  w/o \textsc{mBERT} emb. & 0.0 & 43.0 & 37.7 & 22.9 & 34.7 & 27.5 & 30.6 & 18.1 \\
\hline
\end{tabular}
\end{table}

In general, we can observe as the removal of every component incurs an overall performance degradation, except for the variant without \textsc{mBERT}-based embeddings that obtains a notable gain in accuracy on the graph metrics of the MultiB$_{\text{EU}}$ dataset. More in detail, on the MultiB$_{\text{EU}}$ benchmark, PoS tag embedding ablation has the largest impact in reducing the SF$_1$ score (closely followed by the removal of character and lemma embeddings); and the absence of beam-search decoding is the variant that has less effect in that metric, even improving scores in dependency \added{graph} parsing and holder and expression extraction. While harming full SSA, removing co-parent information seems to benefit sub-tasks such as expression detection and dependency \added{graph} parsing.  Regarding the lack of contextualized word-level embeddings from \textsc{mBERT}, it has a positive impact on the overall accuracy, increasing scores across all metrics except on Targeted F$_1$ and holder identification.

A different pattern can be seen in the DS$_{\text{Unis}}$ dataset. Although lemma embeddings also play an important role; leveraging co-parent features, exploiting beam-search decoding and, especially, employing vector representations extracted from \textsc{mBERT} have a critical impact on the overall model performance (with significant relevance in holder extraction). Moreover, the ablation of character-level embeddings seems to have a positive impact on holder and expression identification as well as in the NSF$_1$ score. Finally, while beam-search decoding is crucial for full SSA, the best Targeted F$_1$ score is achieved by removing it.

Similarly to MultiB$_{\text{EU}}$, we found out in preliminary experiments on the development split of the MultiB$_{\text{CA}}$ benchmark that leveraging \textsc{mBERT}-based embeddings are also counterproductive in full SSA metrics. To further evaluate our approach on test sets, we compare in Table~\ref{tab:bert} our model performance with and without \textsc{mBERT} augmentation. While the impact is less substantial than the one observed in development splits, we can see that leveraging contextualized token-level embeddings
penalizes 
SF$_1$ and Targeted F$_1$ scores
on MultiB$_{\text{EU}}$ and MultiB$_{\text{CA}}$ benchmarks. Since both metrics measure polarity identification accuracy, it might mean that \textsc{mBERT}-based embeddings are harmful for that sub-task on these datasets. \new{A possible explanation is that \textsc{mBERT} was exclusively trained on Basque and Catalan texts from journalistic domains, whereas MultiB$_{\text{EU}}$ and MultiB$_{\text{CA}}$ consist of hotel reviews, a substantially different type of content. This domain mismatch may explain why incorporating \textsc{mBERT}-based embeddings is not beneficial, and may even be detrimental, for polarity detection on these datasets.}

\new{As for the other non-English dataset, \textsc{mBERT} was trained on Norwegian data that, beyond newspaper text, includes small portions of content from government reports, parliamentary transcripts, and informal sources such as blogs. This training data may align better with NoReC$_\text{Fine}$, which comprises reviews from a broad range of domains, including literature, video games, music, various products, films, TV series, and restaurants. Consequently, the use of \textsc{mBERT}-based embeddings contributes to enhanced model performance.}

\new{Regarding the remaining English benchmarks (DS$_{\text{Unis}}$, MPQA and MPQA$_\text{v2}$),  \textsc{mBERT}-based vector representations are crucial for achieving 
good performance 
across all metrics.}

\begin{table}[!t]
\caption{Performance comparison on the test split of our model with and without token-level embeddings extracted from \textsc{mBERT}. We use head-first encoding in all cases.} 
\label{tab:bert}
\centering
\begin{tabular}{@{\hskip 1.0pt}l@{\hskip 1.0pt}c@{\hskip 3.0pt}c@{\hskip 5.0pt}c@{\hskip 5.0pt}c@{\hskip 8.0pt}c@{\hskip 8.0pt}c@{\hskip 5.0pt}c@{\hskip 8.0pt}c@{\hskip 5.0pt}c@{\hskip 1.0pt}}
\hline
 Dataset   &   \textsc{mBERT}         &   \multicolumn{3}{c}{Span F$_1$} &   Targ. &\multicolumn{2}{c}{Dep. G.}  &   \multicolumn{2}{c}{Sent. G.} \\
  &    &   Hold. &   Targ. &   Exp. &  F$_1$ &   UF$_1$ &   LF$_1$ &   NSF$_1$ &   SF$_1$ \\
\hline
\textbf{NoReC$_\text{Fine}$}   & yes &     \textbf{55.8} & \textbf{53.8} & \textbf{52.0} & \textbf{41.4} & \textbf{39.9} & \textbf{32.8} & \textbf{43.6} & \textbf{35.8} \\
& no &   51.1 & 49.9 & 50.2 & 40.1 & 36.5 & 30.7 & 40.0 & 34.3 \\
\hline
\textbf{MultiB$_\text{EU}$}   & yes &   \textbf{67.0} & \textbf{67.7} & \textbf{77.2} & 59.1 & \textbf{68.2} & \textbf{63.4} & \textbf{63.7} & 60.2 \\
& no &  61.4 & \textbf{67.7} & 76.5 & \textbf{61.5} & 65.2 & 61.5 & 63.2 & \textbf{60.5} \\
\hline
\textbf{MultiB$_\text{CA}$}  & yes &    \textbf{54.1} & \textbf{75.4} & \textbf{71.4} & 64.0 & \textbf{68.5} & \textbf{64.4} & \textbf{67.0} & 61.0 \\
& no & 48.9 & 72.9 & 70.3 & \textbf{64.3} & 66.1 & 62.8 & 64.5 & \textbf{61.2} \\
\hline
\textbf{MPQA}  & yes &  \textbf{53.1} & \textbf{57.3} & \textbf{48.5} & \textbf{41.1} & \textbf{47.1} & \textbf{44.6} & \textbf{34.6} & \textbf{27.0}   \\
& no &  35.5 & 42.8 & 37.3 & 31.4 & 32.2 & 29.9 & 24.6 & 20.2 \\
\hline
\textbf{MPQA$_\text{v2}$}  & yes &   \textbf{46.2} & \textbf{49.7} & \textbf{44.7} & \textbf{33.9} & \textbf{40.5} & \textbf{37.6} & \textbf{28.7} & \textbf{22.2} \\
& no &  27.7 & 33.9 & 34.2 & 20.7 & 26.0 & 23.7 & 18.5 & 14.2 \\
\hline
\textbf{DS$_\text{Unis}$} &  yes &   \textbf{40.4} & \textbf{44.3} & \textbf{44.3} & \textbf{32.5} & \textbf{37.3} & \textbf{32.5} & \textbf{35.9} & \textbf{29.9} \\
& no & 11.9 & 34.1 & 38.3 & 22.1 & 32.0 & 26.7 & 27.9 & 21.2 \\
\hline
\end{tabular}
\end{table}

\subsection{Impact of Syntactic Information}
\ext{Alternatively to the head-first and head-final encoding strategies, we further explore the impact of leveraging syntactic information in the encoding process: i.e., instead of choosing the first or last token as head of each node span, a syntactic head is selected based on the information provided by an external dependency parser.\footnote{Syntactic information is already included in the described benchmarks and was provided by SpaCy \citep{ines_montani_2023_10009823} for English, Stanza \citep{qi-etal-2020-stanza} for Basque and Catalan, and UDPipe \citep{straka-strakova-2017-tokenizing} for Norwegian.}
 Table~\ref{tab:syntax} presents the scores obtained by the syntax-based encoding against the head-first and head-final baselines.} 
 
 \ext{We
observe that a syntax-based encoding strategy only 
makes a difference in full SSA for the NoReC$_{\text{Fine}}$ dataset (with an improvement of 4.3 percentage points in SF$_1$ score with respect the other two alternatives). Leveraging syntactic information to encode the nodes of the sentiment graph improves the identification of target and expression nodes, which in turn enhances the overall performance in generating complete sentiment graphs. 
We found no evidence in the data to account for the impact of the syntax-based encoding strategy on the NoReC$_{\text{Fine}}$ dataset compared to the other datasets.}

\ext{
In the MultiB$_\text{EU}$ benchmark we also observe an improvement in the Targeted F$_1$ score, with a gain of 2.6 percentage points over the head-final representation. In this case, the syntax-based strategy proves beneficial for target node identification and thereby increases the Target F$_1$ score, although it does not impact full SSA performance (where the head-final strategy achieves the highest SF$_1$ score). For the remaining datasets, the head-first alternative emerges as the strategy that delivers the best overall performance.
}

\ext{We therefore conclude that, in general, a simple head-first strategy is sufficient for effectively addressing SSA through transition-based dependency graph parsing. The head-first variant appears to produce structures that are easier for the parser to learn, as they are constructed using a transition-based left-to-right method in which the head word is read before being assigned to its dependents. Incorporating additional syntactic information when encoding sentiment graph nodes tends to produce dependency structures that are likely more difficult to construct, and thus leads to accuracy gains only in specific datasets, such as NoReC$_{\text{Fine}}$.}






\begin{table}[!t]
\caption{Accuracy comparison of head-first, head-final and syntax-based encoding strategies.}
\label{tab:syntax}
\centering
\begin{tabular}{@{\hskip 1.0pt}l@{\hskip 1.0pt}l@{\hskip 1.0pt}c@{\hskip 4.0pt}c@{\hskip 4.0pt}c@{\hskip 8.0pt}c@{\hskip 8.0pt}c@{\hskip 4.0pt}c@{\hskip 8.0pt}c@{\hskip 4.0pt}c@{\hskip 1.0pt}}
\hline
 Dataset   &   Encoding         &   \multicolumn{3}{c}{Span F$_1$} &   Targ. &   \multicolumn{2}{c}{Dep. G.} &   \multicolumn{2}{c}{Sen. G.} \\
  &    &   Hold. &   Targ. &   Exp. &  F$_1$ &   UF$_1$ &   LF$_1$ &   NSF$_1$ &   SF$_1$ \\
\hline
\textbf{NoReC$_\text{Fine}$}   & head-first &    55.8 & 53.8 & 52.0 & \textbf{41.4} & 39.9 & 32.8 & 43.6 & 35.8 \\
 & head-final &  \textbf{57.5} & 54.3 & 51.5 & 36.1 & \textbf{45.0} & \textbf{36.5} & 44.2 & 35.8 \\
 & syntax-based & 56.5 & \textbf{55.3} & \textbf{53.2} & 34.9 & 41.5 & 34.8 & \textbf{50.3} & \textbf{40.1} \\
\hline
\textbf{MultiB$_\text{EU}$}  & head-first &    \textbf{67.0} & 67.7 & \textbf{77.2} & 59.1 & \textbf{68.2} & \textbf{63.4} & 63.7 & 60.2 \\
& head-final &    66.6 & 67.9 & 76.6 & 60.5 & 66.9 & 61.4 & \textbf{63.8} & \textbf{60.4} \\
&  syntax-based & 65.7 & \textbf{68.3} & 74.9 & \textbf{63.1} & 60.9 & 57.2 & 63.1 & 59.9 \\
\hline
\textbf{MultiB$_\text{CA}$}  & head-first &    54.1 & \textbf{75.4} & 71.4 & \textbf{64.0} & \textbf{68.5} & \textbf{64.4} & \textbf{67.0} & \textbf{61.0}   \\
& head-final &     \textbf{55.7} & 73.9 & 69.6 & 57.0 & 65.2 & 60.2 & 64.7 & 59.0 \\
& syntax-based & 51.5 & 74.2 & \textbf{71.5} & 59.2 & 67.9 & 63.3 & 66.8 & 60.5 \\
\hline
\textbf{MPQA}  & head-first & \textbf{46.2} & \textbf{49.7} & \textbf{44.7} & \textbf{33.9} & \textbf{40.5} & \textbf{37.6} & \textbf{28.7} & \textbf{22.2} \\
 & head-final & 43.8 & 46.4 & 45.2 & 18.1 & 38.0 & 35.4 & 25.9 & 19.7 \\   
 & syntax-based & 43.2 & 43.4 & 42.5 & 27.4 & 33.9 & 31.0 & 23.9 & 18.6 \\
\hline
\textbf{MPQA$_\text{v2}$}  & head-first &  \textbf{53.1} & \textbf{57.3} & \textbf{48.5} & \textbf{41.1} & \textbf{47.1} & \textbf{44.6} & \textbf{34.6} & \textbf{27.0}   \\
 & head-final &      52.8 & 54.9 & 48.4 & 23.7 & 46.0 & 43.3 & 33.3 & 26.0  \\
 & syntax-based & 47.7 & 54.2 & 48.0 & 24.5 & 45.1 & 41.5 & 31.6 & 23.7 \\
\hline
\textbf{DS$_\text{Unis}$}  & head-first & 40.4 & 44.3 & \textbf{44.3} & \textbf{32.5} & \textbf{37.3} & \textbf{32.5} & \textbf{35.9} & \textbf{29.9} \\
& head-final & \textbf{43.2} & \textbf{44.4} & 43.6 & 28.7 & 36.3 & 30.3 & 32.7 & 25.6 \\
& syntax-based & 11.9 & 34.1 & 38.3 & 22.1 & 32.0 & 26.7 & 27.9 & 21.2 \\
\hline
\end{tabular}
\end{table}

\subsection{Time Complexity}
\label{sec:complexity}
Casting it as dependency graph parsing is one of the most efficient approaches to solve SSA.
Among dependency-based methods, the best-performing graph-based model by \cite{Peng2021} has a $O(n^3)$ worst-case runtime complexity due to the leverage of second-order features. \added{We demonstrate that the transition-based model introduced in this work is more efficient in practice. Although its theoretical running time complexity is $O(n^3)$, the average-case complexity observed in our experiments is $O(n^2)$ over the tested data range.}

\begin{figure*}[h]
\centering
\includegraphics[width=2.5in]{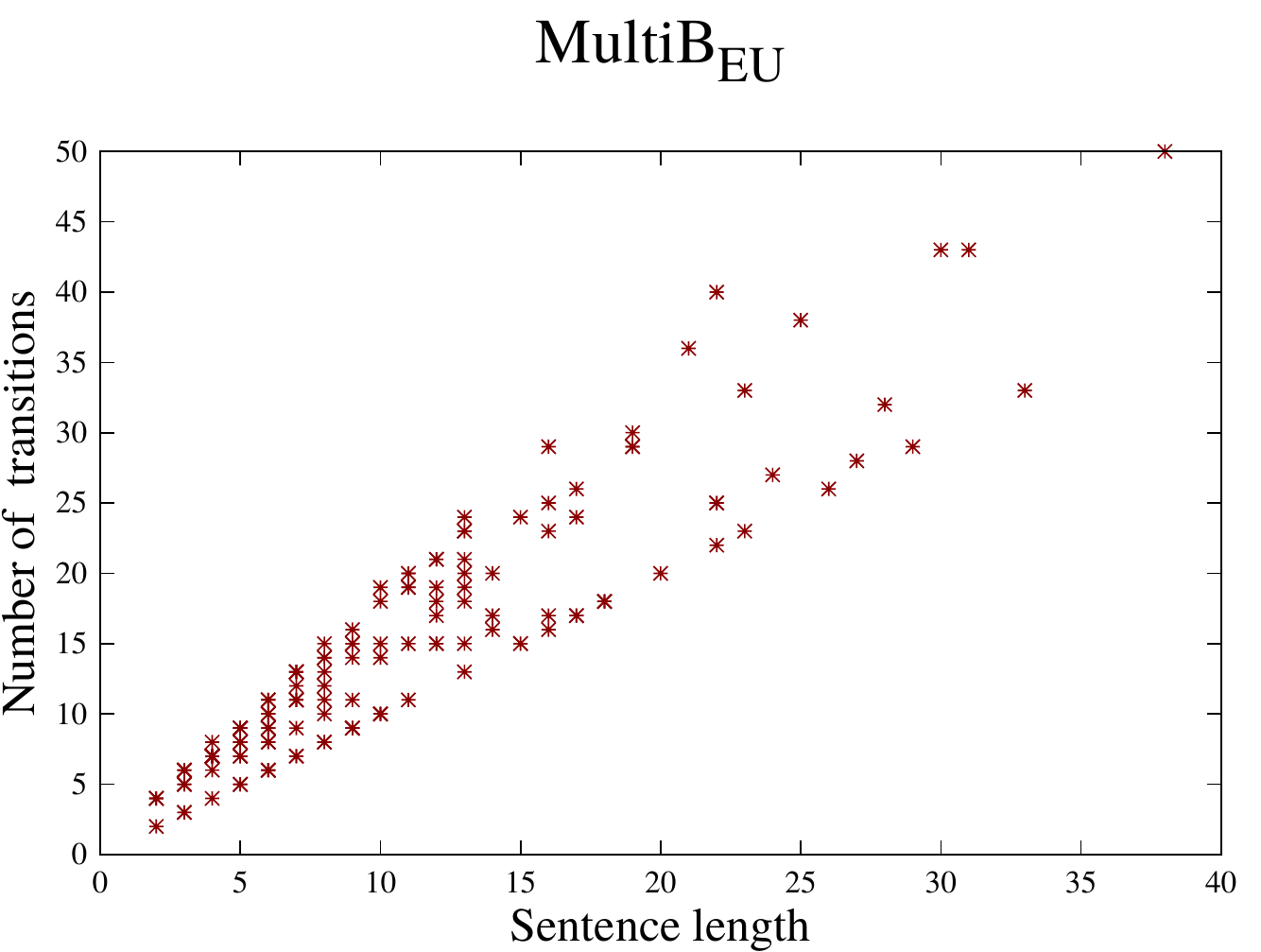}%
\label{fig_first_case}
\hspace{0.01pt}
\includegraphics[width=2.5in]{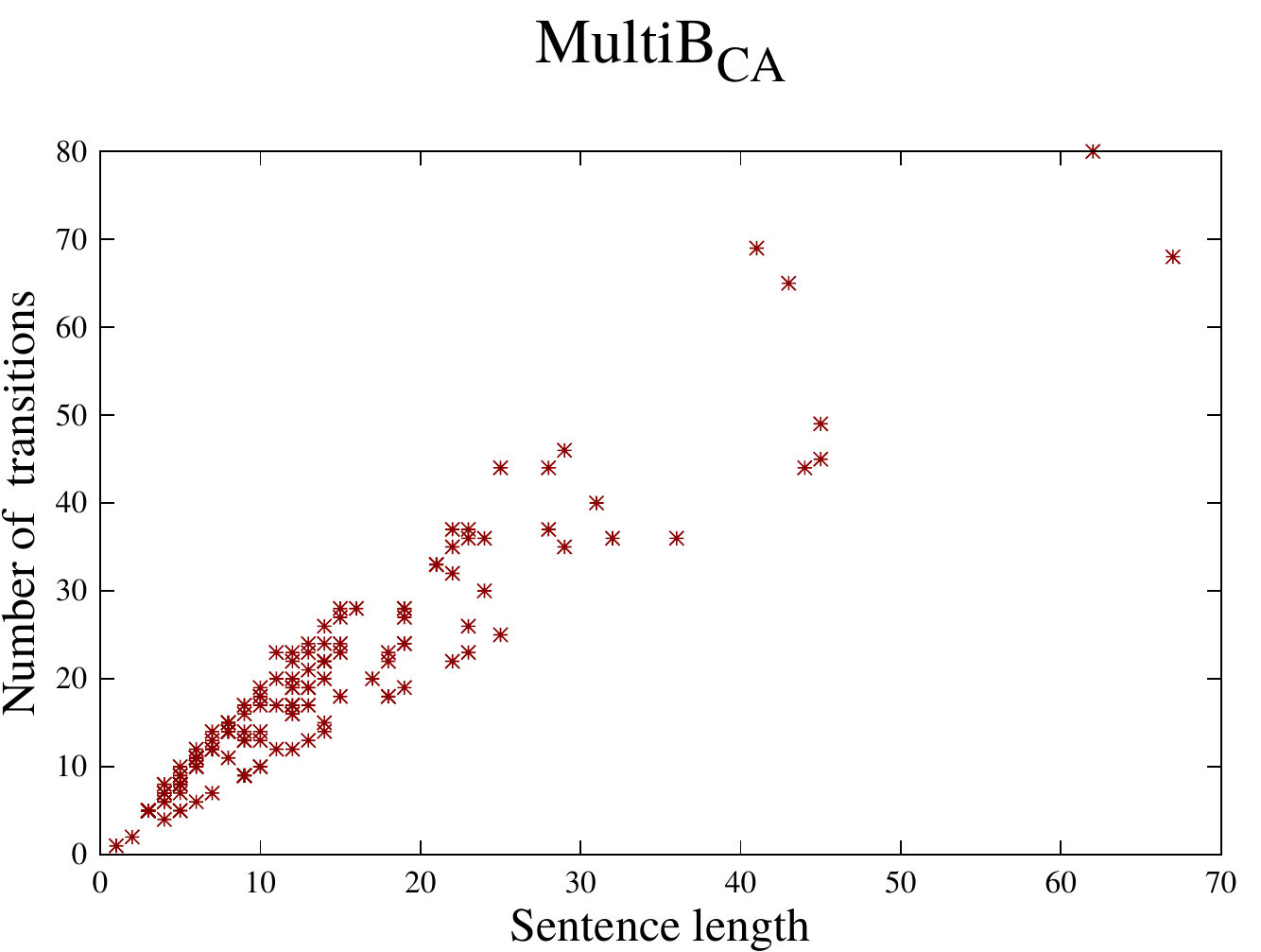}%
\label{fig_second_case}
\hspace{0.01pt}
\includegraphics[width=2.5in]{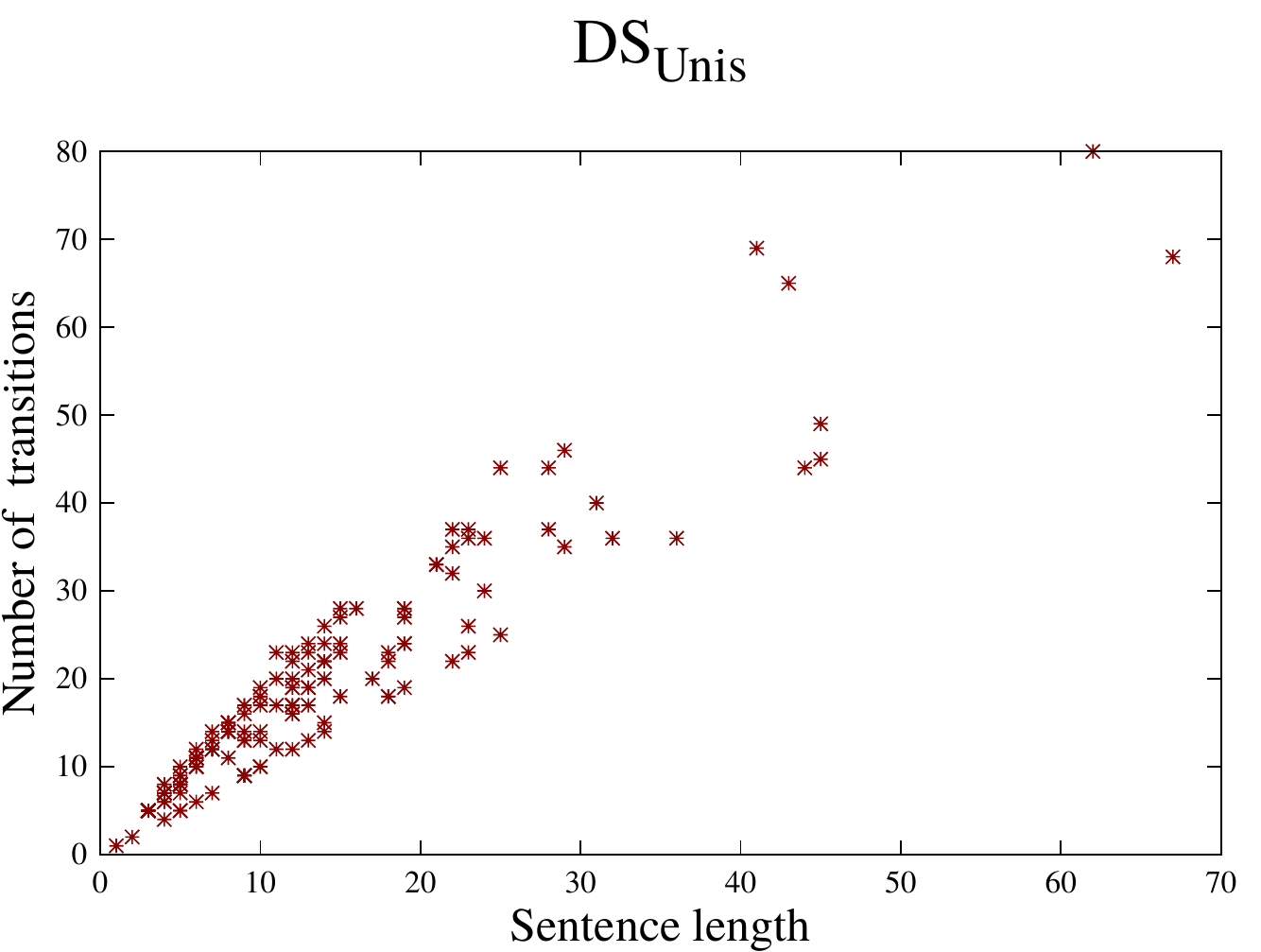}%
\label{fig_3_case}
\hspace{0.01pt}
\includegraphics[width=2.5in]{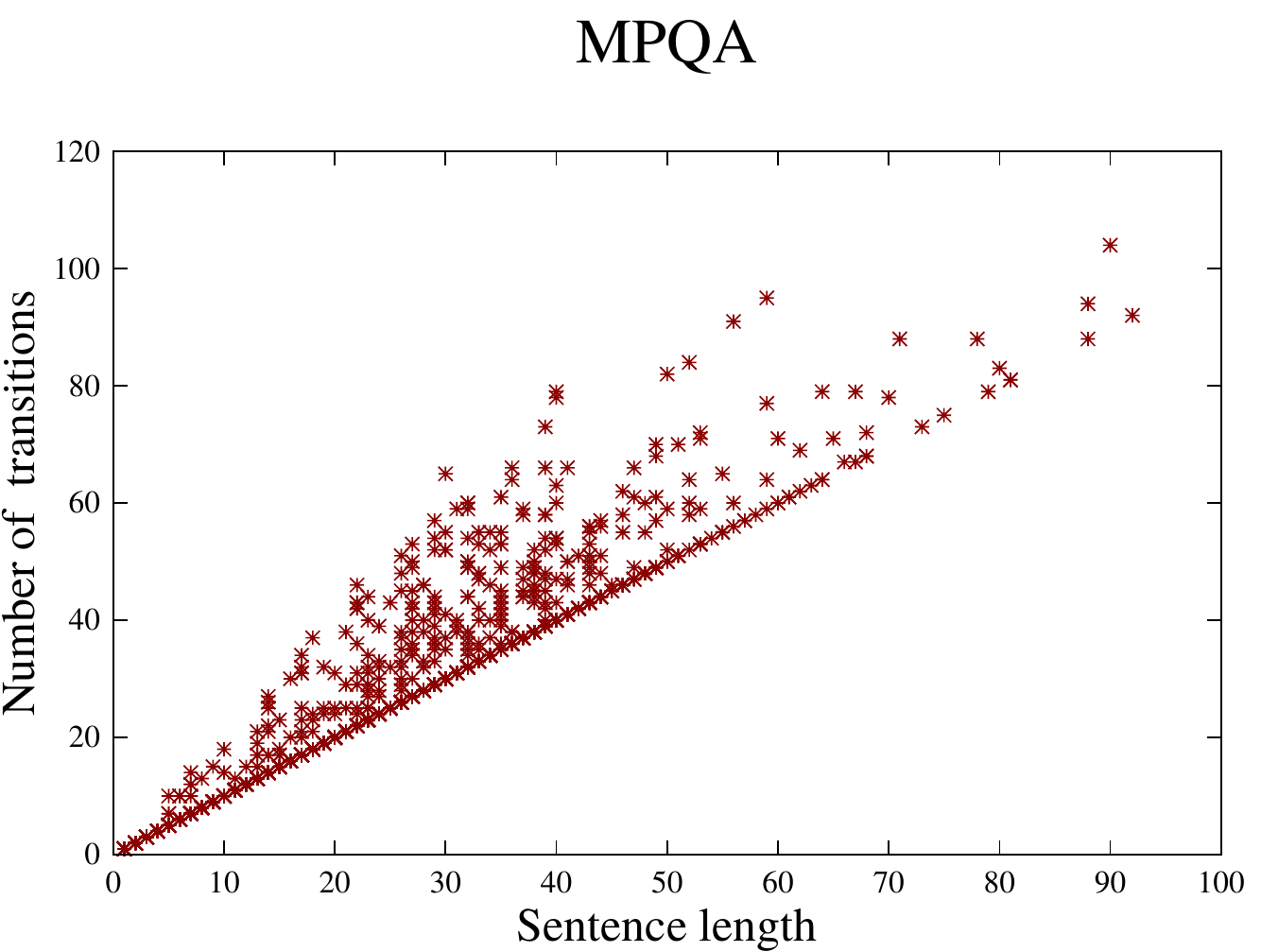}%
\label{fig_4_case}
\hspace{0.01pt}
\includegraphics[width=2.5in]{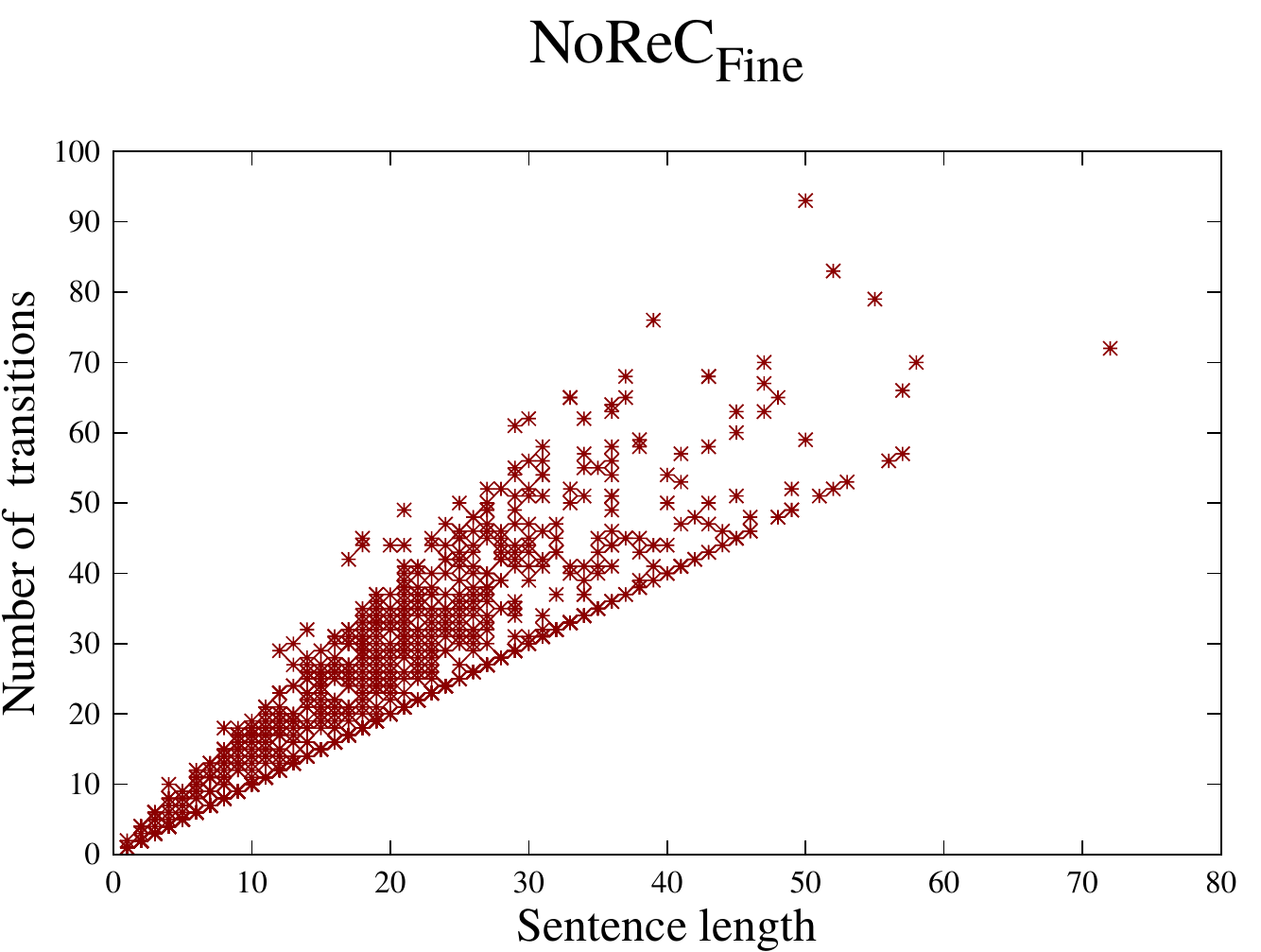}%
\label{fig_5_case}
\caption{Length of transition sequences predicted by our approach relative to the sentence length on development sets for the five SSA benchmarks.}
\label{fig:complexity}
\end{figure*}

Being $n$ the sentence length, a directed graph can potentially include $\Theta(n^2)$ edges at most and, therefore, it would be necessary $O(n^2)$ transitions to build it in the worst case: i.e., $n$ \textsc{Move} actions for reading all input tokens and $n$ \textsc{Attach-to} transitions per token for connecting each one to  all the remaining tokens (excluding itself and including the root node).
Nevertheless, dependency graphs obtained from sentiment graphs in the five SSA benchmarks can be built with just $O(n)$ transitions. To prove this, it has to be determined the parsing complexity of our transition system in practice. Following the work by \cite{Kubler2009}, we study the parsing complexity of our algorithm by examining the relation of the predicted transition sequence length with respect to $n$. In Figure~\ref{fig:complexity}, we graphically present how the number of predicted transitions varies as a function of the number of tokens for every sentence from the development splits of the five SSA datasets. It can be clearly observed a linear relationship across all datasets, meaning that the number of \textsc{Attach-to} actions required per token is substantially low and behaves 
like a constant in the represented linear function. This pattern is supported  
by the fact that
there are notably fewer arcs than tokens in dependency structures 
generated from sentiment graphs of the standard SSA benchmarks. 
In fact, due to the significant amount of unattached tokens, the average ratio of dependency arcs per token in a sentence is less than 1 in all training sets utilized in our experiments: 0.32 in NoReC$_{\text{Fine}}$, 0.52 in MultiB$_{\text{EU}}$, 0.52 in MultiB$_{\text{CA}}$, 0.13 in MPQA and 0.08 in DS$_{\text{Unis}}$. As a consequence, every sentence from SSA datasets
can be parsed in linear time with 2$n$ transitions at most:  $n$ \textsc{Move} transitions plus $n$ \textsc{Attach-to} actions.

Lastly, while our transition system can linearly process input sentences, the described neural architecture requires computing the attention vector $\alpha^t$ over the whole input sentence in $O(n)$ time for each transition prediction. \added{Therefore, although the proposed transition-based approach has a theoretical worst-case time complexity of $O(n^3)$, its average-case complexity on the test data is $O(n^2)$. This performance is comparable to the graph-based baseline of \cite{Barnes2021} and substantially more efficient than the model proposed by \cite{Peng2021}, as well as other task-specific methods that depend on high-order graph-based parsers. }



\section{Conclusion}
\label{sec:conclusions}
Away from the graph-based mainstream, we propose the first transition-based alternative for performing SSA as dependency \added{graph} parsing. We design a transition system that can accurately produce dependency-based representations that, after recovery, will result in well-formed sentiment graphs. In order to implement an efficient SSA approach, we resort to a Pointer Network for action prediction. 
From an extensive evaluation, we show that our transition-based model outperforms previous graph-based methods in practically all standard benchmarks and 
delivers a competitive performance in comparison to task-specific systems.


\new{Moreover, we empirically show that contextualized token-level embeddings from \textsc{mBERT} penalize the accuracy in SSA of our approach on small non-English datasets. We also investigate the impact of syntactic information for encoding sentiment graphs into dependency structures, finding out that syntax-agnostic variants are better options in almost all cases.} \new{Finally, we additionally demonstrate that our transition-based method is one of the most efficient SSA approaches, processing all datasets with a quadratic time-complexity cost in the sentence length.} 

\new{While our approach is robust and accurate, it could certainly benefit from a task-specific fine-tuning of BERT-based embeddings or larger token representations extracted from XLM-R as done by \cite{Samuel2022}. Additionally, an extensive hyperparameter exploration for each specific dataset could also lead to further accuracy gains. However, 
our primary goal is to demonstrate that the proposed approach is effective and practical, even without intensive fine-tuning. This choice highlights the accessibility of our method for real-world scenarios where computational resources may be limited. Moreover, large-scale fine-tuning would require substantial time and resources, which are beyond the scope of our current setup. Instead, we focus on showing the robustness of our model under realistic, resource-constrained conditions.}

\backmatter


\section*{Declarations}

\subsection*{Fundings}
We acknowledge grants SCANNER-UVIGO (PID2020-113230RB-C22) funded by MICIU/AEI/10.13039/501100011033, and LATCHING-UVIGO (PID2023-147129OB-C22) funded by MICIU/AEI/10.13039/501100011033 and ERDF/EU. Open Access funding provided thanks to the CRUE-CSIC agreement with Springer Nature.

\subsection*{Conflict of interest/Competing interests}
The authors declare that they have no known competing financial interests or personal relationships that could have appeared to influence the work reported in this article.

\begin{appendices}
\section{Standard Deviation}
\label{appendix}
In Table~\ref{tab:standdev}, we report standard deviations of scores presented in Tables~\ref{tab:results}, \ref{tab:bert} and \ref{tab:syntax}; and, in Table~\ref{tab:standdevDEV}, we incluide standard deviations of scores in Table~\ref{tab:study}.

\begin{sidewaystable}[!t]
\caption{Standard deviations of scores reported in Tables~\ref{tab:results}, \ref{tab:bert} and \ref{tab:syntax}.}
\label{tab:standdev}
\centering
\begin{tabular}{lclcccccccc}
\hline
 Dataset   &   \textsc{mBERT} & Encoding         &   \multicolumn{3}{c}{Span F$_1$} &   Targeted &   \multicolumn{2}{c}{Dependency Graph} &   \multicolumn{2}{c}{Sent. Graph} \\
  &    &    &   Holder &   Target &   Exp. &  F$_1$ &   UF$_1$ &   LF$_1$ &   NSF$_1$ &   SF$_1$ \\
\hline
\textbf{NoReC$_\text{Fine}$}   & yes & head-first &   5.58 & 1.65 & 1.77 & 1.06 & 1.13 & 0.78 & 0.56 & 0.78 \\
& yes & head-final &   4.41 & 0.65 & 0.42 & 0.67 & 0.73 & 0.49 & 0.46 & 0.64 \\
& yes & syntax-based & 2.40 & 1.04 & 1.58 & 1.44 & 1.56 & 1.14 & 1.40 & 1.06 \\
& no & head-first &   3.24 & 1.38 & 1.15 & 0.91 & 0.93 & 0.42 & 1.10 & 1.11 \\
\hline
\textbf{MultiB$_\text{EU}$}  & yes & head-first &    3.34 & 2.51 & 1.08 & 1.24 & 0.93 & 0.96 & 0.40 & 0.59 \\
& yes & head-final &   2.29 & 1.07 & 1.16 & 0.92 & 1.04 & 0.87 & 0.87 & 0.46 \\
& yes & syntax-based & 2.64 & 1.45 & 0.60 & 0.91 & 0.74 & 0.65 & 0.68 & 0.88 \\
& no & head-first &  3.42 & 0.75 & 0.44 & 0.83 & 1.00 & 0.90 & 0.57 & 0.38 \\
\hline
\textbf{MultiB$_\text{CA}$}   & yes & head-first &  3.16 & 1.25 & 1.20 & 1.46 & 1.36 & 0.88 & 1.02 & 0.36 \\
& yes & head-final &     4.51 & 0.70 & 0.98 & 1.23 & 0.79 & 0.33 & 0.64 & 0.51 \\
& yes & syntax-based & 3.16 & 0.43 & 0.84 & 1.73 & 1.01 & 0.97 & 0.42 & 0.26 \\
& no & head-first & 1.30 & 1.18 & 0.82 & 0.72 & 0.54 & 0.45 & 0.55 & 0.81 \\
\hline
\textbf{MPQA}   & yes & head-first &  1.54 & 1.66 & 1.21 & 1.24 & 2.00 & 1.89 & 1.09 & 0.96 \\
& yes & head-final &      1.59 & 0.95 & 0.79 & 0.35 & 1.15 & 1.07 & 0.63 & 0.53 \\
& yes & syntax-based & 1.49 & 1.98 & 0.22 & 2.15 & 0.60 & 0.43 & 0.58 & 0.37 \\
& no & head-first &   1.11 & 2.10 & 0.73 & 1.01 & 1.67 & 1.62 & 0.96 & 1.12 \\
\hline
\textbf{MPQA$_\text{v2}$}   & yes & head-first & 1.72 & 1.20 & 0.75 & 1.69 & 0.30 & 0.38 & 1.78 & 1.49 \\
& yes & head-final &      3.28 & 1.54 & 1.99 & 1.45 & 0.63 & 0.50 & 1.96 & 1.95 \\
& yes & syntax-based & 1.50 & 1.55 & 1.64 & 1.70 & 1.53 & 1.41 & 1.71 & 1.18 \\
& no & head-first &  1.35 & 0.70 & 2.10 & 1.21 & 0.88 & 0.97 & 1.13 & 1.09 \\
\hline
\textbf{DS$_\text{Unis}$} &   yes & head-first &    3.94 & 3.88 & 1.88 & 1.62 & 2.14 & 1.39 & 3.51 & 1.32 \\
& yes & head-final &  7.24 & 3.85 & 3.45 & 2.21 & 2.49 & 1.97 & 4.07 & 3.36 \\
& yes & syntax-based & 6.63 & 2.74 & 2.96 & 2.29 & 2.27 & 2.57 & 1.83 & 3.31 \\
& no & head-first & 6.66 & 3.76 & 2.21 & 2.14 & 1.59 & 1.53 & 3.85 & 2.50 \\
\hline
\end{tabular}
\end{sidewaystable}

\begin{sidewaystable}[!t]
\caption{Standard deviations of scores reported in Table~\ref{tab:study}.}
\label{tab:standdevDEV}
\centering
\begin{small}
\begin{tabular}{llcccccccc}
\hline
 Dataset   &   \textsc{Model}         &   \multicolumn{3}{c}{Span F$_1$} &   Targeted &   \multicolumn{2}{c}{Depen. Graph} &   \multicolumn{2}{c}{Sentim. Graph} \\
  &    &   Holder &   Target &   Exp. &  F$_1$ &   UF$_1$ &   LF$_1$ &   NSF$_1$ &   SF$_1$ \\
\hline
\textbf{MultiB$_\text{EU}$}   & full &     5.13 & 1.13 & 1.89 & 0.85 & 1.25 & 0.87 & 2.00 & 1.69 \\
& w/o PoS tag emb. & 2.87 & 1.47 & 2.00 & 2.33 & 1.10 & 1.10 & 0.94 & 0.97 \\
& w/o character emb. & 3.66 & 1.07 & 0.74 & 1.26 & 1.10 & 1.37 & 1.44 & 0.90 \\
& w/o lemma emb. & 4.73 & 1.82 & 1.24 & 1.79 & 0.94 & 1.18 & 1.94 & 1.36 \\
& w/o co-parent feat. & 4.32 & 1.94 & 1.27 & 2.07 & 1.27 & 0.60 & 1.13 & 1.16 \\
& w/o beam search & 5.70 & 1.30 & 0.84 & 1.73 & 0.83 & 0.62 & 1.52 & 0.97 \\
& w/o \textsc{mBERT} emb. & 4.80 & 1.50 & 0.95 & 2.34 & 0.76 & 1.01 & 2.49 & 1.25 \\
\hline
\textbf{DS$_\text{Unis}$} & full &   0.98 & 0.36 & 0.58 & 1.57 & 1.03 & 1.03 & 0.18 & 0.94 \\
& w/o PoS tag emb. & 9.73 & 3.19 & 2.24 & 1.83 & 2.18 & 1.30 & 1.53 & 1.68 \\
& w/o character emb. &  13.05 & 4.69 & 1.23 & 3.69 & 2.29 & 2.14 & 4.05 & 3.39 \\
& w/o lemma emb. & 9.64 & 1.21 & 2.06 & 3.00 & 2.15 & 2.67 & 1.90 & 3.44 \\
& w/o co-parent feat. & 10.95 & 1.46 & 2.75 & 5.11 & 2.27 & 2.53 & 2.25 & 2.84 \\ 
& w/o beam search & 9.83 & 1.43 & 1.45 & 3.58 & 0.94 & 1.34 & 1.91 & 2.38 \\
& w/o \textsc{mBERT} emb. & 0.00 & 3.45 & 1.72 & 2.42 & 1.26 & 1.30 & 0.36 & 1.27 \\
\hline
\end{tabular}
\end{small}
\end{sidewaystable}

\end{appendices}

\bibliography{anthology,main}

\end{document}